\documentclass[sigconf,nonacm]{acmart} 
\usepackage{amsmath}
 
\usepackage{amssymb}
\usepackage{amsfonts}
\usepackage{algorithm}
\usepackage{algpseudocode}
\usepackage{multirow}
\usepackage{booktabs}
\usepackage[most]{tcolorbox}

\usepackage{xcolor}

\tcbset{
    acmstyle/.style={
        enhanced,              
        colback=gray!5!white,  
        colframe=black!75!white,
        coltitle=white,        
        fonttitle=\bfseries,   
        arc=1mm,               
        boxrule=0.8pt,         
        left=2mm, right=2mm, top=2mm, bottom=2mm, 
        attach boxed title to top center={yshift=-2mm, xshift=0mm}, 
        boxed title style={colback=black!75!white} 
    }
}

\newtcolorbox{promptbox}[2][]{
    acmstyle,                  
    title={#2},                
    #1                         
}

\newtcolorbox{definitionbox}[2][]{
    acmstyle,                  
    title={#2},
    #1
}

\usepackage{tikz}
\usepackage{placeins}

\begin{document}

\title{Boundary-Aware NL2SQL: Integrating Reliability through Hybrid Reward and Data Synthesis}

\author{Songsong Tian}
\authornote{These authors contributed equally to this work as co-first authors.}
\email{tiansongsong@lixiang.com}
\affiliation{%
  \institution{Li Auto Inc.}
  \streetaddress{No.4 Hengxing Road, Gaoliying Town, Shunyi District}
  \city{Beijing}
  \country{China}}

\author{Kongsheng Zhuo}
\authornotemark[1]
\email{zhuokongsheng@lixiang.com}
\affiliation{%
  \institution{Li Auto Inc.}
  \streetaddress{No.4 Hengxing Road, Gaoliying Town, Shunyi District}
  \city{Beijing}
  \country{China}}

\author{Zhendong Wang}
\authornotemark[1]
\email{yuchenlogin@bupt.edu.cn}
\affiliation{%
  \institution{Beijing University of Posts and Telecommunications}
  \streetaddress{10 Xitucheng Rd, Beitaipingzhuang, Haidian District}
  \city{Beijing}
  \country{China}}

\author{Rong Shen}
\authornote{These authors contributed equally to this work as co-second authors.}
\email{shenrong@lixiang.com}
\affiliation{%
  \institution{Li Auto Inc.}
  \streetaddress{No.4 Hengxing Road, Gaoliying Town, Shunyi District}
  \city{Beijing}
  \country{China}}

\author{Shengtao Zhang}
\authornotemark[2]
\authornote{Corresponding author.}
\email{zhangshengtao@lixiang.com}
\affiliation{%
  \institution{Li Auto Inc.}
  \streetaddress{No.4 Hengxing Road, Gaoliying Town, Shunyi District}
  \city{Beijing}
  \country{China}}

\author{Yong Wu}
\authornotemark[2]
\email{wuyong5@lixiang.com}
\affiliation{%
  \institution{Li Auto Inc.}
  \streetaddress{No.4 Hengxing Road, Gaoliying Town, Shunyi District}
  \city{Beijing}
  \country{China}}
\begin{abstract}
Natural Language to SQL (NL2SQL) systems powered by Large Language Models (LLMs) have shown strong performance on academic benchmarks, yet their effectiveness degrades substantially in enterprise environments with large schemas and complex business logic. Empirical studies on real-world production databases indicate that state-of-the-art NL2SQL systems achieve limited execution accuracy on realistic business queries. More critically, when queries are ambiguous or unanswerable, models tend to \textit{force an answer}, producing plausible but semantically incorrect SQL rather than abstaining or requesting clarification. These limitations in both accuracy and boundary-aware behavior hinder the reliable deployment of NL2SQL systems in enterprise settings.

In this paper, we present BAR-SQL (\textbf{B}oundary-\textbf{A}ware \textbf{R}eliable NL2SQL), a unified training framework that embeds reliability and boundary awareness directly into the generation process. We introduce a Seed Mutation data synthesis paradigm that constructs a representative enterprise corpus, explicitly encompassing multi-step analytical queries alongside boundary cases including ambiguity and schema limitations. To ensure interpretability, we employ Knowledge-Grounded Reasoning Synthesis, which produces Chain-of-Thought traces explicitly anchored in schema metadata and business rules. The model is trained through a two-stage process: Supervised Fine-Tuning (SFT) followed by Reinforcement Learning via Group Relative Policy Optimization. We design a Task-Conditioned Hybrid Reward mechanism that simultaneously optimizes SQL execution accuracy—leveraging Abstract Syntax Tree analysis and dense result matching—and semantic precision in abstention responses.
To evaluate reliability alongside generation accuracy, we construct and release Ent-SQL-Bench, which jointly assesses SQL precision and boundary-aware abstention across ambiguous and unanswerable queries. Experimental results on this benchmark demonstrate that BAR-SQL achieves 91.48\% average accuracy, outperforming leading proprietary models, including Claude 4.5 Sonnet and GPT-5, in both SQL generation quality and boundary-aware abstention capability. The source code and benchmark are available anonymously at: \url{https://github.com/TianSongS/BAR-SQL}.

\end{abstract}

\keywords{NL2SQL,
Data Synthesis,
Reward Design,·
Enterprise Databases,
Knowledge-Grounded Reasoning,
Enterprise Business Intelligence}

\maketitle

\section{Introduction}
Natural Language to SQL (NL2SQL) has become a key interface for democratizing data access, allowing non-experts to query relational databases without learning SQL \cite{zhang2024sqlfuse, liu2025survey}. Early systems predominantly relied on rule-based parsers or task-specific neural architectures. The advent of Large Language Models (LLMs), however, has dramatically reshaped the landscape\cite{sadga-cai2021sadga, hwang2019comprehensive, liu2025survey}. With few-shot prompting and instruction tuning, modern LLM-based methods capture user intent and database schemas far more effectively \cite{dailsql-gao2023text, liu2025survey}. Performance on standard cross-domain benchmarks has increased accordingly: execution accuracy on Spider 1.0 has advanced from approximately 53.5\% in 2020 to over 88.1\% by 2025 \cite{zhong2020grounded,Sql-r1-ma2025sql}, suggesting strong performance on well-defined schemas and unambiguous queries.
However, evaluations on real enterprise databases with genuine user queries show that this performance does not reliably carry over to production settings, even with common enhancements such as schema grounding and retrieval augmentation~\cite{stonebraker2025dbos}. 

The recent Spider 2.0 benchmark, which assembles 632 real-world data-analysis workflow problems, reveals a significant discrepancy between academic scores and production reality \cite{spider2-lei2024spider}. Agent-based methods dominate these benchmarks (e.g., ByteBrain-Agent at 84.10\% execution accuracy) but typically rely on multiple LLM calls and iterative feedback, leading to high latency and complex architectures. FDABench \cite{wang2025fdabench} shows that general-purpose data agents incur 178–1,015 seconds of inference time per query, and multi-agent or reflection setups may require up to 31.1 external model calls without commensurate accuracy gains. Moreover, many leading agent systems remain closed-source, limiting reproducibility, customization, and scalable integration in cost-sensitive enterprise environments.

Building robust, enterprise-ready NL2SQL systems is further constrained by the lack of high-quality, domain-specific training data. Creating such datasets requires domain experts to manually author and validate tens of thousands of \textit{(Question, SQL)} pairs that accurately encode complex and evolving business rules and database schemas. For most organizations—particularly small and medium-sized enterprises (SMEs)—this manual curation is expensive and time-intensive \cite{liu2025survey}. This highlights the need for high-fidelity, automated data synthesis pipelines that can generate domain-tailored training corpora with enterprise-grade quality.

In addition to reasoning complexity and data scarcity, query ambiguity often leads to intent uncertainty and downstream intent misalignment, posing a pervasive challenge for NL2SQL. Real-world users frequently submit underspecified requests such as “top customers,” which admit multiple valid interpretations (e.g., by revenue, order volume, or other metrics), each requiring a different SQL query.
Standard LLM-based NL2SQL systems often exhibit forced answering: even when intent is ambiguous or the question is unanswerable given the schema, they still generate syntactically valid SQL by committing to an arbitrary interpretation, resulting in executable yet semantically misaligned queries \cite{ding2025ambisql}.
However, production Business Intelligence (BI) systems require robust reliability alongside high execution accuracy. This includes the capacity to recognize knowledge boundaries, request clarification for missing intent, and abstain from issuing unsupported queries. 
While prior work has attempted to address reliability through post-hoc mechanisms such as conformal prediction and heuristic abstention \cite{chen2025reliable}, these methods operate as external filters atop a frozen generator, making boundary awareness and clarification heuristic add-ons rather than intrinsic modeling capabilities.

To address these interrelated challenges, we propose BAR-SQL, a unified framework that internalizes both reliability and complex reasoning as \textit{first-class} training objectives. The BAR-SQL framework commences with a cold-start phase grounded in a \textit{Seed-Mutation Synthesis paradigm}, which constructs a high-fidelity corpus by applying controlled logical and structural mutations to business-grounded seeds. To ensure interpretability and faithful reasoning, we introduce \textit{Knowledge-Grounded Reasoning Synthesis} (KGRS), which generates evidence-based Chain-of-Thought (CoT) traces explicitly anchored in schema metadata, metric definitions, and domain rules. Building upon this corpus, we perform unified generative training using Group Relative Policy Optimization (GRPO) guided by a Task-Condition-Aware Hybrid Reward (TCHR). This reward mechanism jointly optimizes SQL execution accuracy and boundary-aware abstention, explicitly penalizing hallucinated SQL in unanswerable scenarios while rewarding precise, executable queries and appropriate clarification or rejection responses. This design aligns the model with the reliability imperatives of production BI environments.

In summary, our primary contributions are as follows:
\begin{itemize}
\item \textbf{Seed-Mutation Synthesis Paradigm and Open-Source Benchmark}: We introduce a scalable data synthesis pipeline that evolves seed scenarios anchored in business logic into comprehensive enterprise datasets via systematic mutations. Leveraging this paradigm, we construct and release Ent-SQL-Bench to rigorously evaluate the execution accuracy of complex SQL generation and the reliability of abstention mechanisms in the face of ambiguous or unanswerable queries.

\item \textbf{Knowledge-Grounded Reasoning Synthesis (KGRS)}: We propose a universal framework for synthesizing evidence-based reasoning traces. By anchoring CoT generation in schema metadata and business rules, KGRS reinforces the causal relationship between business logic and model outputs, ensuring that reasoning remains faithful to underlying data constraints.

\item \textbf{Task-Condition-Aware Hybrid Reward (TCHR)}: We propose a hybrid reward mechanism that jointly optimizes SQL execution accuracy and boundary-aware abstention. TCHR integrates fine-grained feedback from AST-based structural similarity, dense execution result matching, and semantic embedding consistency, enabling the model to learn both precise SQL generation and reliable refusal behavior within a unified reinforcement learning framework.

\item \textbf{Unified Reinforcement Learning Architecture}: We present a comprehensive training pipeline that unifies the proposed synthesis and reward mechanisms under a GRPO-based policy optimization strategy. This architecture addresses the alignment challenge between strict syntax adherence and ambiguous boundary detection. Experimental results demonstrate that our approach achieves 91.48\% average accuracy, exhibiting superior reliability compared to leading proprietary LLMs in enterprise-grade contexts.


\end{itemize}

\section{Data Synthesis and Benchmark Construction}
\label{sec:data_synthesis}
We construct the training corpus and benchmark through a systematic three-stage framework. First, we generate a high-fidelity \textit{seed dataset} to serve as the structural foundation (Section~\ref{sec:seed_generation}). Building upon these seeds, we apply controlled transformations to synthesize \textit{seed-based mutations}, evolving the data from standard queries into advanced scenarios that require complex reasoning and interactive boundary awareness (Section~\ref{sec:7moreTasks}). Finally, based on this corpus, we curate Ent-SQL-Bench (Section~\ref{sec:benchmark_desc}), a benchmark explicitly designed to evaluate both the precision and reliability of enterprise NL2SQL models.

\subsection{Seed Data Generation}
\label{sec:seed_generation}

We define a \textbf{seed} as a high-fidelity, unambiguous NL-SQL pair that strictly conforms to the underlying database schema and business definitions. Serving as the structural backbone of our framework, the seed set represents the \textit{Standard SQL} category and provides the foundational semantics for all subsequent evolutionary mutations.

\begin{figure*}[htbp]
\centering
\includegraphics[width=1.05\textwidth]{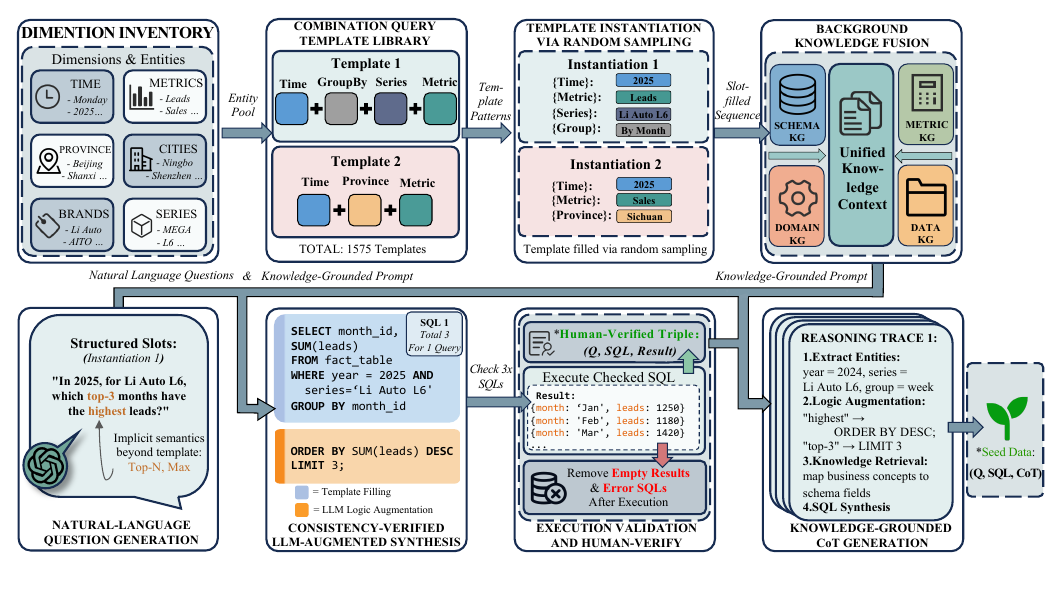} 
\caption{Seed data synthesis pipeline. The workflow begins with a dimension inventory containing dimensions and entities (TIME, METRICS, PROVINCE, CITIES, BRANDS, SERIES), which are used to construct 1,575 combinatorial query templates. Templates are instantiated via random sampling, then transformed into natural-language questions through LLM generation with Domain-KG-constrained analytical augmentation (e.g., Top-K ranking, extremum operations), ensuring all augmentations reference only schema-defined entities. The system performs consistency-verified LLM-augmented SQL synthesis by fusing background knowledge context (Schema KG, Metric KG, Domain KG, and Data KG). Generated SQLs undergo execution validation and human verification to produce human-verified triplets \texttt{(Q, SQL, Result)}, eliminating empty results and erroneous queries. Finally, knowledge-grounded Chain-of-Thought (CoT) reasoning traces are synthesized through a 4-step process (Extract Entities, Logic Augment, Knowledge Retrieval, SQL Synthesis) to produce validated seed data triplets \texttt{(Q, SQL, CoT)}.}
\label{fig:seed_generation_diagram}
\end{figure*}

To guarantee correctness and eliminate hallucination-induced inconsistencies during SQL generation, we design a rigorously constrained Structured Seed Generation pipeline. This pipeline reflects a practitioner-oriented approach, with the complete workflow illustrated in Figure~\ref{fig:seed_generation_diagram}.

\subsubsection{Dimension Inventory and Semantic Repository.}
We first construct a dimension repository containing 30 business dimensions and 832 fine-grained entities spanning temporal entities (143 time expressions), business metrics (111 indicators), geographical attributes (34 provinces and 370 cities), brand names (62), and vehicle series (112).

Each entity is mapped to a deterministic SQL fragment, thereby forming a precise semantic space that links natural-language expressions to executable conditions.


\subsubsection{Combinatorial Query Template Library.}
Leveraging the dimension repository, we construct a library of 1,575 query templates. Each template encodes a structural analytic pattern such as \texttt{[Time, Brand, Metric, GroupBy]} or \texttt{[Province, City, Time, VehicleType, Metric]}.
These templates enable controlled composition of realistic enterprise BI queries,
incorporating multi-dimensional filters and analytical grouping logic. Each template can be instantiated with concrete values across business dimensions such as time, region, product, and channel, combined with different group-by configurations. A single template can thus generate various query shapes that mirror typical BI workloads, slicing metrics for specific segments and aggregating them under various reporting views.

\subsubsection{Template Instantiation via Random Sampling.}
For each template, entities are sampled either uniformly or according to business priors. For instance, instantiating the pattern \texttt{[Time, Series, Metric, GroupBy]} may yield the structured slot-filled sequence:

\noindent
\texttt{[Year 2025, Li Auto L6, Leads, By Month]}
This procedure guarantees semantic coherence while enabling large-scale
data generation.

\subsubsection{Background Knowledge Fusion for SQL Grounding.}
To ensure deterministic SQL generation and NL question generation, we provide the model with a unified knowledge context comprising:
\begin{itemize}
\item \textbf{Schema KG:} Complete DDL definitions of fact and dimension tables, including all keys and semantic fields.
\item \textbf{Metric KG:} Definitions, aliases, mandatory constraints, and computation logic (e.g., ``Net New Leads'' requires \texttt{is\_net\_leads = 1} and \texttt{count(distinct customer\_account\_id)}).
\item \textbf{Domain KG:} Rules for interpreting time ranges such as ``last 7 days'', grouping defaults, and business conventions including default channel granularity and distinctions between vehicle-type fields.
\item \textbf{Data KG:} Mappings such as channel hierarchies (e.g., ``Outreach'' mapped to \texttt{first\_leads\_channel\_tag\_name}).
\end{itemize}
This fusion aims to ensure that SQL generation is grounded not only in schema structure but also in business constraints, metric semantics, and domain-specific rules.

\subsubsection{Natural-Language Question Generation with Emergent Augmentation.}
A high-capacity LLM transforms the slot-filled structure into fluent, human-style queries. LLM is provided with the complete background knowledge and permitted to introduce analytical variations naturally fitting the business context. For instance, the slot-filled structure ``Year 2025, Li Auto L6, Leads, By Month'' may be rendered as any of the following:

\noindent
\textit{What is the monthly lead count for Li Auto L6 in 2025?} 
(direct translation)

\noindent
\textit{In 2025, which month had the highest leads for Li Auto L6?} 
(extremum selection)

\noindent
\textit{Show me the top-3 months by lead count for Li Auto L6 in 2025.} 
(ranking query)


\noindent
We do not enforce a fixed distribution of augmentation types. Instead, the Knowledge Context inherently constrains the generation space: Schema KG limits referenceable entities, Metric KG defines valid aggregation semantics, and Domain KG specifies permissible analytical operations, while subsequent validation stages ensure correctness.

\subsubsection{Consistency-Verified LLM-Augmented SQL Synthesis.}
To ensure structural determinism and eliminate invalid queries caused by 
random mutations or enhancements in the previous stage, we employ a 
multi-sample consistency verification mechanism. Specifically, for each 
question generated in the previous stage, the LLM independently generates 
three SQL query samples under identical conditions. Only when all three 
samples are completely consistent is the SQL query accepted and allowed 
to proceed to the next stage, thereby guaranteeing the structural 
determinism of the seed set.

\subsubsection{Execution Validation and Human Verify.}
Each accepted SQL is executed against a production-like environment. Queries yielding empty results, execution errors, or business-rule violations are discarded. A stratified subset undergoes manual inspection to ensure correctness and fidelity.

\subsubsection{Knowledge-Grounded CoT Generation.}
For each validated $(Question, SQL)$ pair, we generate a knowledge-grounded CoT trace that explicates the derivation of the SQL. To ensure that reasoning remains concise and tightly coupled to domain knowledge, we introduce a unified synthesis framework KGRS. It conditions the teacher model on the structured knowledge context, the question, and the validated SQL, subsequently reconstructing a compact reasoning trace that cites the relevant knowledge employed for metric mapping, temporal inference, and filter selection.
This process yields consistent, evidence-based CoTs aligned with the information actually required to produce the target SQL.

\begin{definitionbox}{Knowledge-Grounded Reasoning Synthesis (KGRS)}
We propose KGRS, a unified paradigm designed to synthesize interpretable reasoning traces by anchoring the generation process in structured domain knowledge.
Formally, given a knowledge context $\mathcal{C}$ (encompassing schema, metric definitions, and domain rules), an input state $\mathcal{Q}$ such as a user question or error message, and a validated target $\mathcal{T}$ such as Gold SQL, clarification request, or rejection, KGRS employs a teacher LLM to reconstruct the logical rationale $\mathcal{R}$ that bridges $\mathcal{Q}$ and $\mathcal{T}$:
    \begin{equation}
    \mathcal{R} \sim P_{\text{LLM}}(\cdot \mid \mathcal{C}, \mathcal{Q}, \mathcal{T}; \mathcal{E}_{\text{static}})
    \end{equation}
where $\mathcal{E}_{\text{static}}$ denotes a small, fixed set of human-annotated exemplars used to standardize the reasoning format.
Unlike standard inference, KGRS functions as a \textit{reverse-reasoning} mechanism, ensuring that the synthesized CoT explicitly cites evidence from $\mathcal{C}$ to justify the target $\mathcal{T}$. This paradigm is applied universally across subsequent tasks—from \textit{Standard SQL} generation to other further tasks related to interactive boundary awareness—thereby equipping the student model with robust, evidence-based reasoning capabilities.
\end{definitionbox}


\subsection{Seed-Based Derivation for Advanced Tasks}
\label{sec:7moreTasks}
To transcend simple information retrieval, we enhance the seed data through logical complication and error simulation. Crucially, for all evolved samples, we leverage the unified reasoning synthesis paradigm defined in KGRS to generate evidence-based CoT traces. This ensures that even for complex or negative tasks, the model's output is strictly grounded in the provided Schema and Knowledge contexts.

\subsubsection{Multi-Step Reasoning}
Complex business queries frequently require nested logic or window functions. We demonstrate the evolution from seed data to \textit{Multi-Step Reasoning} through a three-stage pipeline:
\begin{enumerate}
    \item \textbf{Generation:} Based on a specified difficulty level, we rewrite a seed question to incorporate multi-table joins or trend analysis (e.g., evolving a sales query into ``How many percentage points did the leads share increase in 2024 compared to 2023?'').
    \item \textbf{Validation:} We generate the corresponding complex SQL and validate its execution consistency.
    \item \textbf{KGRS Explanation:} We treat the complex SQL as the target $\mathcal{T}$ and the evolved question as $\mathcal{Q}$. Employing KGRS, the teacher model reconstructs the decomposition logic based on the schema context $\mathcal{C}$. As evidenced in synthesis logs, the synthesized CoT explicitly cites: ``I will use CTEs to calculate 2023 and 2024 statistics separately...'' before deriving the final join operation.
\end{enumerate}

\subsubsection{Reflection}
To endow the model with debugging capabilities, we synthesize \textit{Reflection} data. We deliberately inject syntactic or logical errors into valid SQL queries and execute them to capture authentic error messages.
Here, the KGRS input $\mathcal{Q}$ is defined as the tuple $(Question, SQL_{error}, ErrorMsg)$, and the target $\mathcal{T}$ is $SQL_{correct}$. The reasoning trace is synthesized as a ``debugging log,'' where the system analyzes the error message against the schema definition to identify the root cause and derive the corrected SQL.

\subsubsection{Degenerate Dimension}
To ensure robustness against schema evolution, we synthesize \textit{Degenerate Dimension} data. We physically merge dimension attributes into fact tables within the prompt's schema definition (thereby modifying the context $\mathcal{C}$) and rewrite the target SQL $\mathcal{T}$ to eliminate \texttt{JOIN} operations.
The reasoning synthesis is then re-executed with this modified context. The teacher model is compelled to generate a new reasoning path that selects fields directly from the fact table (e.g., \texttt{dwd\_sale\_target.province\_name}) rather than joining \texttt{dim\_area}, thereby training the student model to rely on the current schema definition rather than memorized patterns.

\textbf{The above three categories enhance SQL seed data.} A distinctive feature of our approach is the transformation of the model into an active collaborator. We extend the KGRS paradigm to non-SQL targets, wherein the validated target $\mathcal{T}$ comprises a natural language response (clarification, follow-up, or rejection). KGRS synthesizes the logical justification for why SQL generation should be suspended. \textbf{The following are four data categories that embody interactive boundary awareness for non-SQL targets.}

\subsubsection{Ambiguity Clarification.}
Real-world queries frequently contain vague terms. We employ an \textit{Ambiguity Injection} mechanism wherein precise seed queries are rewritten into ambiguous variants (e.g., ``Inner Mongolia'' to ``northern province''; specific date to ``recently'').
The target $\mathcal{T}$ is formulated as a clarification question: ``Could you please specify the time range and indicator?''. KGRS synthesizes the detection logic: the CoT cites schema evidence to demonstrate that no field matches ``indicators'' and that ``recently'' is not a defined time window in the Domain KG, thus justifying the clarification request.


\subsubsection{Constraint Follow-Up.}
Business metrics often imply mandatory constraints defined in the Knowledge Graph. We synthesize scenarios wherein the user's query $\mathcal{Q}$ fails to satisfy a constraint specified in the Metric KG (part of context $\mathcal{C}$).
The target $\mathcal{T}$ is formulated as a specific inquiry requesting the missing parameter mandated by the metric's definition. The teacher model subsequently generates a reasoning trace that explicitly contrasts the query against the Metric KG in $\mathcal{C}$, identifies the violation of the mandatory filter constraint (e.g., ``Net New Leads'' requires ``Car Series''), and justifies the suspension of SQL generation in favor of user intervention.

\subsubsection{Knowledge Rejection.}
To mitigate hallucinations, we synthesize rejection data by deliberately introducing knowledge deficits into the prompt context $\mathcal{C}$.
\begin{enumerate}
    \item \textit{Dimension Rejection:} We remove specific dimension definitions (e.g., deleting \texttt{series\_name}) from the schema.
    \item \textit{Metric Rejection:} We query for undefined metrics such as ``High-Intent Leads''.
\end{enumerate}
In these cases, the target $\mathcal{T}$ is a refusal message. The synthesis process searches the tampered context $\mathcal{C}$ and records the failure. The resulting CoT explicitly documents the search-and-failure process, providing a grounded rationale for the rejection rather than a generic refusal.

\subsection{Enterprise Reliability Benchmark}
\label{sec:benchmark_desc}

Existing benchmarks, such as Spider 2.0 and BIRD, predominantly assess the \textit{capability upper-bound} of NL2SQL models by focusing on the complexity of queries that models can successfully resolve. However, they largely overlook the \textit{safety lower-bound} critical to enterprise environments, wherein forced answering (hallucinating SQL for invalid queries) poses greater risk than appropriate refusal.

To address this gap, we construct a specialized evaluation benchmark: Ent-SQL-Bench, explicitly designed to assess \textbf{Reliability} alongside \textbf{Capability}. Derived through stratified sampling from our mutated corpus, this testbed comprises 1,262 high-quality instances that remain strictly held-out during training. Distinct from standard test sets, it features a balanced distribution between:

\begin{itemize}
    \item Positive Tasks ($\approx$50\%): Scenarios wherein valid SQL generation is expected, challenging the model's reasoning depth (e.g., \textit{Multi-Step Reasoning}, \textit{Degenerate Dimension}).
    \item Negative \& Interactive Tasks ($\approx$50\%): Scenarios designed to elicit potential failure modes, requiring the model to demonstrate boundary awareness by rejecting unanswerable queries or requesting necessary clarifications.
\end{itemize}

This dual-track design establishes a rigorous evaluation standard that penalizes blind instruction-following, compelling agents to function not merely as code translators but as boundary-aware data analysts.

\section{Model Training and Reward Design}
\label{sec:method}
\begin{figure}
\centering
\includegraphics[width=\linewidth]{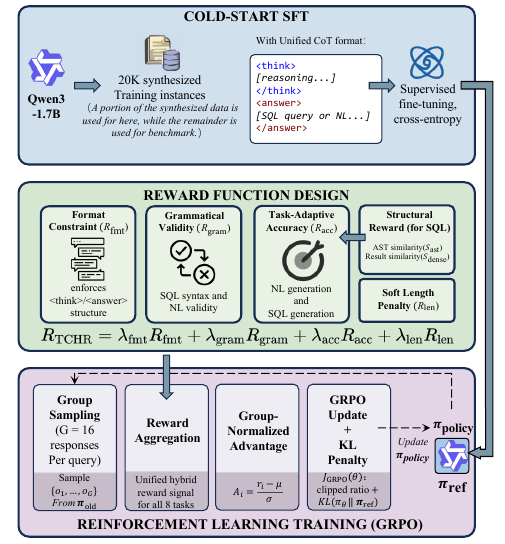}
\caption{Overview of the BAR-SQL training pipeline. The framework consists of three stages: cold-start SFT with unified CoT format, TCHR reward function design (integrating Format, Grammar, Accuracy, and Length components), and GRPO-based policy optimization. The TCHR serves as the core feedback mechanism guiding the model to handle heterogeneous NL2SQL behaviors.}
\label{fig:ModelTrain}
\end{figure}

We employ BAR-SQL, a multi-stage optimization framework that integrates supervised learning and reinforcement learning to align the model with diverse NL2SQL behaviors. The model is first initialized through cold-start supervised fine-tuning, and subsequently optimized via GRPO using a unified hybrid reward function. The overall training pipeline is illustrated in Figure~\ref{fig:ModelTrain}.

\subsection{Cold Start}
\label{sec:cold-start}

We initialize the model through a SFT cold-start stage prior to reinforcement learning.
This setup aligns with the multi-stage training paradigm adopted in recent large-scale reasoning models such as DeepSeek-R1~\cite{guo2025deepseek}, wherein initial exposure to high-quality structured supervision precedes policy optimization.

Specifically, we synthesized approximately \textbf{20K training instances}, covering the seven data categories described in section~\ref{sec:7moreTasks}, plus the \textit{Standard SQL} category (eight in total).



All instances adhere to a unified reasoning–response format: 
\texttt{<think> ... </think> <answer> ... </answer>}, 
where the \texttt{<think>} block contains the reasoning trace 
and the \texttt{<answer>} block contains the final output.

Given the strict latency, memory, and cost constraints in enterprise on-premise deployment, we deliberately initialize the policy model with Qwen3-1.7B~\cite{qwen3technicalreport}, a lightweight instruction-tuned model, which is fine-tuned for 5 epochs using standard cross-entropy loss to obtain a stable initial policy for reinforcement learning, reducing exploration instability while maintaining structured and readable outputs.

\subsection{Reinforcement Learning Training}

Building upon the cold-started model, we further optimize the policy GRPO~\cite{guo2025deepseek} to better align the model with the demands of enterprise NL2SQL tasks.
GRPO estimates advantages from within-group reward statistics and therefore avoids the need for a separate critic network.

Formally, for each user query \(q\), we sample a group of \(G=16\) responses \(\{o_1, \dots, o_G\}\) from the current policy \(\pi_{\theta_{\text{old}}}\).
Each response \(o_i\) is assigned a scalar reward \(r_i = R_{\text{TCHR}}(q, o_i, \tau)\) using a task-condition-aware hybrid reward signal, whose specific formulation is detailed in the following subsection.
The advantage is computed by normalizing rewards within the group:
\begin{equation}
A_i = \frac{r_i - \mu(\{r_1,\dots,r_G\})}{\sigma(\{r_1,\dots,r_G\})}.
\label{eq:advantage}
\end{equation}

The policy is updated by maximizing a clipped surrogate objective with a KL-divergence penalty:

\begin{equation}
\begin{aligned}
\mathcal{J}_{\text{GRPO}}(\theta) = &  \mathbb{E} \Bigg[ \frac{1}{G} \sum_{i=1}^G \Bigg(  \min\Big( \frac{\pi_\theta(o_i|q)}{\pi_{\theta_{\text{old}}}(o_i|q)} A_i, 
 \text{clip}\Big( \frac{\pi_\theta(o_i|q)}{\pi_{\theta_{\text{old}}}(o_i|q)}, \\
& 1-\epsilon, 1+\epsilon \Big) A_i \Bigg) 
 - \beta \, \mathbb{D}_{\text{KL}}\big( \pi_\theta \| \pi_{\text{ref}} \big) \Bigg) \Bigg],
\end{aligned}
\label{eq:grpo}
\end{equation}

where \(\pi_{\text{ref}}\) denotes the cold-started reference model, \(\epsilon=0.2\), and \(\beta=0.01\).
The KL penalty constrains the policy from deviating excessively from the domain knowledge and output structure acquired during supervised fine-tuning.

Compared to standard GRPO setups, the utilization of a unified hybrid reward signal enables a single policy to be optimized across heterogeneous NL2SQL behaviors, encompassing SQL generation, clarification elicitation, and constraint validation.
Consequently, the optimized policy preserves the structured reasoning behavior established during the cold-start phase while adapting its outputs to diverse enterprise query scenarios under task-specific reward constraints.

\subsection{Reward Function Formulation}
\label{sec:method-reward}
\begin{figure*}[htbp]
\centering
\includegraphics[width=1.0\textwidth]{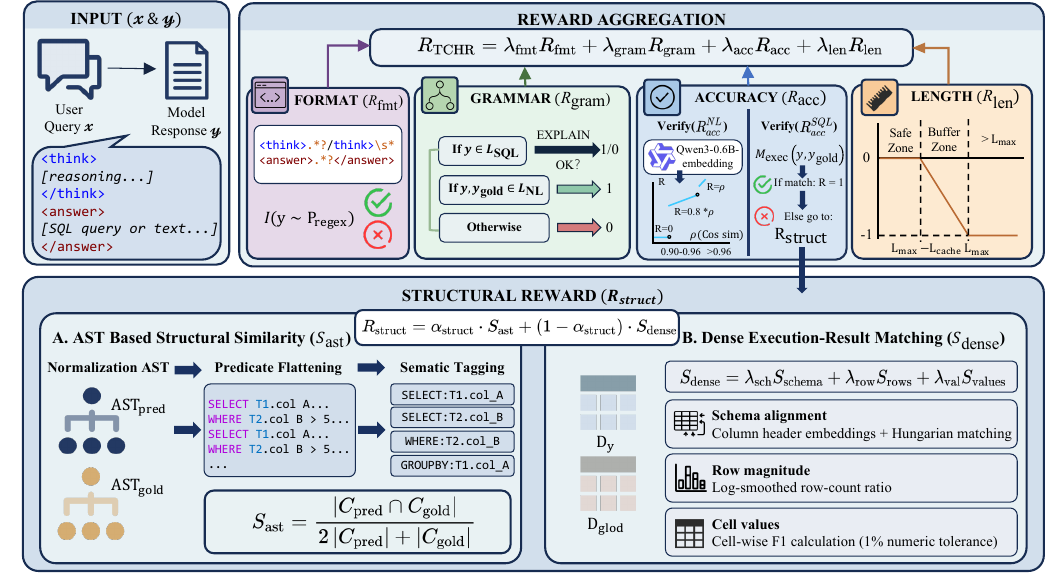} 
\caption{Composition of the Task-Condition-Aware Hybrid Reward (TCHR) function. The reward aggregates four components—Format, Grammar, Accuracy, and Length—with task-adaptive weights. The Accuracy component bifurcates into Natural Language verification (using embedding-based similarity) and SQL verification (using execution matching with structural reward fallback). The structural reward combines AST-based similarity and dense execution-result matching to provide granular feedback when binary execution fails.}
\label{fig:reward_composition}
\end{figure*}

To align the model with the multifaceted requirements of NL2SQL, which span precise SQL generation, semantic disambiguation, and interactive boundary awareness, we design the \textbf{Task-Condition-Aware Hybrid Reward (TCHR)}.This reward function is a weighted combination of four components: Format Constraint (\( R_{\text{fmt}} \)), Grammatical Validity (\( R_{\text{gram}} \)), Adaptive Accuracy (\( R_{\text{acc}} \)), and a Soft Length Penalty (\( R_{\text{len}} \)). Formally, for an input query \( x \) and generated response \( y \), \( R_{\text{TCHR}} \) is defined as:
\begin{equation}
\label{eq:reward}
    R_{\text{TCHR}}(x, y) = \lambda_{\text{acc}} R_{\text{acc}} + \lambda_{\text{fmt}} R_{\text{fmt}} + \lambda_{\text{gram}} R_{\text{gram}} + \lambda_{\text{len}} R_{\text{len}}
\end{equation}
Here, \( \lambda_{\text{acc}}, \lambda_{\text{fmt}}, \lambda_{\text{gram}}, \lambda_{\text{len}} \) are weighting coefficients that balance the contribution of each reward component.  
An overview of the reward function composition is illustrated in Figure~\ref{fig:reward_composition}.

\subsubsection{Format Constraint ($R_{\text{fmt}}$)}
To enforce the CoT reasoning process, we require the model to structure its output using specific XML-style tags. Let $\mathcal{P}_{\text{regex}}$ denote the pattern \texttt{<think>.*?</think>\\s*<answer>.*?</answer>}. The reward is binary:
\begin{equation}
    R_{\text{fmt}}(y) = \mathbb{I}(y \sim \mathcal{P}_{\text{regex}})
\end{equation}
where $\mathbb{I}(\cdot)$ denotes the indicator function. This constraint ensures that the reasoning trace and the final answer can be reliably parsed.

\subsubsection{Grammatical Validity ($R_{\text{gram}}$)}
This component verifies the executability of the generated output $y$. Let $\Phi_{\text{syn}}(y)$ denote the boolean result of the syntax check using the database engine's \texttt{EXPLAIN} command. To formally categorize the output types, let $\mathcal{L}_{\text{SQL}}$ represent the set of SQL statements and $\mathcal{L}_{\text{NL}}$ represent the set of natural language responses. The reward function is defined as:
\begin{equation}
    R_{\text{gram}}(y, y_{\text{gold}}) = 
    \begin{cases} 
        \mathbb{I}(\Phi_{\text{syn}}(y)) & \text{if } y \in \mathcal{L}_{\text{SQL}} \\
        1 & \text{if } y, y_{\text{gold}} \in \mathcal{L}_{\text{NL}} \\
        0 & \text{otherwise}
    \end{cases}
\end{equation}
where $\mathbb{I}(\cdot)$ is the indicator function. The first case rewards syntactically valid SQL generations. The second case assigns full credit when both the prediction and the ground truth are natural language (e.g., for unanswerable queries), assuming intrinsic grammatical validity for non-executable text. The zero score applies to format mismatches or invalid type combinations. This segmented reward structure effectively encourages the model to correctly distinguish between NL tasks and executable SQL tasks based on the prompt and target output type.

\subsubsection{Soft Length Penalty ($R_{\text{len}}$)}
To encourage concise reasoning without truncating valid steps, we adopt a soft length penalty inspired by the overlong reward shaping technique employed in DAPO \cite{yu2025dapo}. This component penalizes responses approaching the predefined generation limit, thereby discouraging verbosity while preserving necessary reasoning steps. Let \( L(y) \) denote the token length of the generated sequence, \( L_{\max} \) the maximum allowed generation length, and \( L_{\text{cache}} \) a buffer margin. The penalty \( R_{\text{len}}(y) \) is defined as:

\begin{equation}
R_{\text{len}}(y) =
\begin{cases}
0, & L(y) \leq L_{\max} - L_{\text{cache}} \\[6pt]
\dfrac{(L_{\max} - L_{\text{cache}}) - L(y)}{L_{\text{cache}}}, & L_{\max} - L_{\text{cache}} < L(y) \leq L_{\max} \\[6pt]
-1, & L(y) > L_{\max}
\end{cases}
\end{equation}

The penalty remains zero within the safe length, increases linearly within the buffer zone, and applies a strong negative signal when the limit is exceeded. In practice, this mechanism not only manages context window utilization but also contributes to output diversity. We observe that it mitigates token-level repetition, an issue widely reported in the community when employing models such as the Qwen3 series for reasoning tasks.

\subsubsection{Task-Adaptive Accuracy ($R_{\text{acc}}$)}
\label{subsec:task-adaptive-accuracy}

To handle the heterogeneous requirements of our dataset, we divide tasks into two modes: natural language (NL) generation and SQL generation. Accordingly, the accuracy reward $R_{\text{acc}}$ is defined separately for these two modalities.

\textbf{1. NL Verification}
This branch covers tasks whose target outputs are textual responses rather than executable code, including four categories: \textit{Ambiguity Clarification}, \textit{Constraint Follow-Up}, \textit{Dimension Rejection}, and \textit{Metric Rejection}.

For natural-language responses, exact string matching is too rigid to capture semantic equivalence. While a dedicated reward model could be trained to assess response quality, this approach is computationally expensive. We instead adopt embedding-based similarity as a lightweight yet effective proxy for semantic matching. Empirical results confirm that this method balances efficiency with robust alignment to human judgment. Concretely, we compute cosine similarity \(\rho\) between embeddings of the generated response \(y\) and the ground truth \(y_{\text{gold}}\) using the Qwen3-0.6B-embedding model.
To filter semantic noise, we apply a discrete step function:

\begin{equation}
R_{\text{acc}}^{\text{NL}} =
\begin{cases}
\rho & \text{if } \rho \ge 0.96, \\
0.8 \times \rho & \text{if } 0.90 \le \rho < 0.96, \\
0.0 & \text{otherwise}.
\end{cases}
\label{eq:r-acc-nl}
\end{equation}

\textbf{2. SQL Generation Verification}
This branch covers tasks requiring executable SQL outputs: \textit{Standard SQL}, \textit{Multi-Step Reasoning}, \textit{Reflection}, and \textit{Degenerate Dimension}.

The primary reward is binary execution correctness. 

Let $M_{\mathrm{exec}}(y, y_{\mathrm{gold}})$ be an indicator that equals 1.0 
if the execution result sets are identical and 0 otherwise. 
To provide dense feedback when execution fails (or when binary rewards 
are too sparse), we introduce a fallback structural reward component 
$R_{\text{struct}}$:

\begin{equation}
R_{\text{acc}}^{\text{SQL}} =
\begin{cases}
1.0 & \text{if } M_{\text{exec}}(y, y_{\text{gold}}) = 1, \\
R_{\text{struct}}(y, y_{\text{gold}}) & \text{otherwise}.
\end{cases}
\label{eq:r-acc-sql}
\end{equation}

\paragraph{Structural Reward ($R_{\text{struct}}$).}

Triggered when strict execution matching fails, $R_{\text{struct}}$ provides dense feedback by combining syntactic fidelity with a soft approximation of execution accuracy. It guides the model toward valid SQL structures even when the final result set is incorrect. We compute it as a weighted sum of an AST-based structural similarity score $S_{\text{ast}}$ and a dense execution-oriented score $S_{\text{dense}}$, balanced by a hyperparameter $\alpha_{\text{struct}} \in [0, 1]$:
\begin{equation}
R_{\text{struct}} =
\alpha_{\text{struct}} \cdot S_{\text{ast}}(y, y_{\text{gold}})
+ (1 - \alpha_{\text{struct}}) \cdot S_{\text{dense}}(D_{y}, D_{\text{gold}}).
\label{eq:r-struct}
\end{equation}

\paragraph{Structural Similarity Metric ($S_{\text{ast}}$).}

Standard string metrics do not capture the order invariance of SQL syntax (e.g., \texttt{WHERE A AND B} $\equiv$ \texttt{WHERE B AND A}). We therefore implement an AST-flattening pipeline that projects the hierarchical query structure into an order-independent set of atomic components. The pipeline consists of three stages:
\begin{enumerate}
    \item \textbf{Normalization}: convert identifiers to lowercase and strip quote characters to obtain a canonical representation.
    \item \textbf{Predicate Flattening}: flatten recursive logical structures (e.g., nested \texttt{AND}/\texttt{OR} trees) into linear lists to handle commutativity.
    \item \textbf{Semantic Tagging}: prepend each component with a context-aware tag to distinguish identical tokens in different clauses (e.g., a column in \texttt{SELECT} vs. \texttt{GROUP BY}).
\end{enumerate}

Let $\mathcal{C}$ denote the set of normalized, flattened, and tagged components. We define $S_{\text{ast}}$ as the F1-score between the predicted set $\mathcal{C}_{\text{pred}}$ and the gold set $\mathcal{C}_{\text{gold}}$:
\begin{equation}
S_{\text{ast}} =
\frac{2 \cdot \lvert \mathcal{C}_{\text{pred}} \cap \mathcal{C}_{\text{gold}} \rvert}
{\lvert \mathcal{C}_{\text{pred}} \rvert + \lvert \mathcal{C}_{\text{gold}} \rvert}.
\label{eq:s-ast}
\end{equation}

\paragraph{Execution Result Matching ($S_{\text{dense}}$).}

For complex business queries, binary rewards can be excessively sparse. Since partial data retrieval may still reflect partial reasoning success, we evaluate the generated result DataFrame $D_{y}$ against the gold DataFrame $D_{\text{gold}}$. The metric approximates correctness along three dimensions: schema alignment, row-count magnitude, and cell-value precision:
\begin{equation}
S_{\text{dense}} =
\lambda_{\text{sch}} \cdot S_{\text{schema}}
+ \lambda_{\text{row}} \cdot S_{\text{rows}}
+ \lambda_{\text{val}} \cdot S_{\text{values}},
\label{eq:s-dense}
\end{equation}
where $\lambda_{\text{sch}}, \lambda_{\text{row}}, \lambda_{\text{val}}$ are hyperparameters.

Specifically:
\begin{itemize}
    \item $S_{\text{schema}}$ uses an embedding model to obtain vector representations of column headers for both $D_{y}$ and $D_{\text{gold}}$, then applies the Hungarian algorithm for optimal semantic alignment.
    \item $S_{\text{rows}}$ applies logarithmic smoothing to the ratio of row counts to penalize large magnitude discrepancies.
    \item $S_{\text{values}}$ computes a cell-wise F1-score with a 1\% numerical tolerance to account for floating-point variation.
\end{itemize}

\section{Experiments}
\subsection{Setup}

\textbf{Metrics.} Given the hybrid nature of our tasks, ranging from deterministic code generation to semantic reasoning, we employ a dual-metric evaluation protocol:
\begin{itemize}
    \item \textbf{Execution Accuracy ($\text{Acc}_{\text{exec}}$):} For tasks yielding executable SQL, we measure execution accuracy. A generated query is deemed correct if its execution result set strictly matches that of the ground truth SQL on the database engine.
    \item \textbf{Semantic Consistency ($\text{Acc}_\text{sem}$):} For tasks whose target output comprises natural language, such as \textit{Ambiguity Clarification}, we adopt an LLM-as-a-Judge approach. We employ GPT-4o as the evaluator to assess semantic consistency between the model output and the ground truth, with majority voting over $k=3$ independent judgments per sample.
\end{itemize}

\textbf{Implementation Settings.} We utilize Qwen3 as the backbone model. The training pipeline is implemented using the \texttt{ms-swift} framework (v3.9.3) and consists of two phases:
\begin{itemize}
    \item \textbf{SFT:} We perform full-parameter fine-tuning on the synthesized dataset for 10 epochs. The training is configured with a learning rate of $5 \times 10^{-5}$ and a warmup ratio of 0.05. We set the per-device batch size to 2 with 16 gradient accumulation steps. DeepSpeed ZeRO-2 is employed for memory optimization.
    \item \textbf{GRPO Alignment:} Following SFT, we apply GRPO to align the model with the hybrid reward function defined in Eq.~\ref{eq:reward}. 
    We set the learning rate to $1 \times 10^{-6}$, the group size (number of generations per prompt) to $G = 16$. The reward weights are set to $\lambda_{\text{acc}} = 1.5$, $\lambda_{\text{fmt}} = 1.0$, $\lambda_{\text{gram}} = 0.9$, $\lambda_{\text{len}} = 0.8$, and $\alpha_{\text{struct}}=0.5$. To ensure efficient exploration, the sampling temperature is set to 1.0.

\end{itemize}

\textbf{Environment.} All experiments are conducted on a high performance computing cluster equipped with NVIDIA A800-SXM4-80GB GPUs. The distributed training infrastructure utilizes 48 GPUs (6 nodes $\times$ 8 GPUs) for the training processes and a separate node with 8 GPUs dedicated to vLLM-based rollout generation to accelerate the RLHF data collection. The software stack comprises PyTorch 2.6.0, DeepSpeed 0.18.2, and CUDA 12.4, employing Flash Attention for memory optimization.

\subsection{Main Results}
\label{sec:main_results}

Table~\ref{tab:main_results} presents a comprehensive comparison between BAR-SQL and several state-of-the-art LLMs. Empowered by GRPO alignment, BAR-SQL achieves an average accuracy of 91.48\%, outperforming the strongest baseline competitor (Claude-4.5-Sonnet) by a substantial margin of 43.46\%.

\begin{table*}[t]
\centering
\small
\caption{
Performance comparison across SQL generation tasks and boundary interaction tasks.
Dim. Deg. denotes \textit{Degenerate Dimension}, and Dim. Rej. denotes \textit{Dimension Rejection}.
The category \textit{General-Purpose Baselines} includes both proprietary (e.g., GPT-5) and open-source (e.g., DeepSeek-V3.2) models.
}
\label{tab:main_results}
\resizebox{\textwidth}{!}{
\begin{tabular}{lccccccccc}
\toprule
\multirow{2}{*}{\textbf{Model}} & \multicolumn{4}{c}{\textbf{SQL Generation Tasks}} & \multicolumn{4}{c}{\textbf{Boundary Interaction Tasks}} & \multirow{2}{*}{\textbf{Avg.}} \\
\cmidrule(lr){2-5} \cmidrule(lr){6-9}
 & \textbf{Dim. Deg.} & \textbf{Reflection} & \textbf{Std. SQL} & \textbf{Multi-Step} & \textbf{Ambiguity} & \textbf{Dim. Rej.} & \textbf{Metric Rej.} & \textbf{Follow-Up} & \\
\midrule
\textit{General-Purpose Baselines} & & & & & & & & & \\
Claude-4.5-Sonnet & 44.07\% & 69.79\% & 60.56\% & 69.72\% & 16.13\% & 18.64\% & 87.50\% & 17.75\% & 48.02\% \\
Gemini3-Pro & 49.15\% & 64.52\% & 52.69\% & 62.50\% & 0.00\% & 0.00\% & 10.34\% & 0.00\% & 29.90\% \\
GPT-4o & 40.68\% & 28.13\% & 22.54\% & 17.96\% & 2.15\% & 0.00\% & 34.38\% & 2.90\% & 18.59\% \\
GPT-5 & 35.59\% & 80.21\% & 46.48\% & 67.61\% & 5.56\% & 13.79\% & 70.49\% & 48.24\% & 46.01\% \\
DeepSeek-V3.2 & 42.37\% & 82.29\% & 47.89\% & 41.90\% & 0.00\% & 1.69\% & 17.19\% & 0.00\% & 29.17\% \\
\midrule
\textit{Ours} & & & & & & & & & \\
Qwen3-1.7B (Base) & 10.17\% & 38.54\% & 14.08\% & 4.93\% & 0.00\% & 0.00\% & 0.00\% & 0.00\% & 8.47\% \\
Qwen3-1.7B (SFT) & 77.97\% & 83.33\% & 62.32\% & 63.73\% & 68.75\% & 42.37\% & 86.12\% & 55.80\% & 67.55\% \\
\textbf{BAR-SQL} & \textbf{93.22\%} & \textbf{93.75\%} & \textbf{90.75\%} & \textbf{81.69\%} & \textbf{94.51\%} & \textbf{93.31\%} & \textbf{92.27\%} & \textbf{92.36\%} & \textbf{91.48\%} \\
\bottomrule
\end{tabular}
}
\end{table*}

\textbf{Analysis of General-Purpose Baselines: The Hallucination Dilemma.}
A critical observation reveals a common limitation shared by general-purpose LLMs, regardless of whether they are proprietary or open-weights. While these models demonstrate competent performance in passive tasks such as \textit{Standard SQL} generation (Claude-4.5: 60.56\%) and \textit{Reflection} (DeepSeek-V3.2: 82.29\%), they exhibit severe deficiencies in boundary interaction tasks. Specifically, in tasks requiring \textit{Ambiguity Clarification} and \textit{Dimension Rejection}, they achieve low scores. For instance, Gemini3-Pro and DeepSeek-V3.2 score nearly 0\% in ambiguity detection.
This indicates that without domain-specific alignment, general LLMs struggle to recognize knowledge boundaries. Rather than seeking clarification or rejecting invalid queries, they tend to hallucinate plausible-looking yet factually incorrect SQL queries (e.g., fabricating columns for missing dimensions). This forced answering behavior renders them unreliable for direct enterprise BI deployment, despite their general coding proficiency.

\textbf{Efficacy of the Seed-Mutation Paradigm.}
The base Qwen3-1.7B model, lacking domain exposure, struggles significantly with these tasks, achieving an average accuracy of only 8.47\%. However, SFT using our synthesized corpus significantly elevates performance to 67.55\%, validating the quality of our data synthesis methodology described in Section~\ref{sec:data_synthesis}. The SFT model learns to handle SQL and metric verification effectively but continues to struggle with subtle boundary cases such as \textit{Dimension Rejection} (42.37\%) and \textit{Constraint Follow-Up} (55.80\%).

\textbf{The Impact of GRPO Alignment.}
The introduction of GRPO provides the decisive performance leap, elevating average accuracy from 67.55\% to 91.48\%. The most substantial gains are observed in tasks requiring strict adherence to business logic. For instance, \textit{Metric Rejection} and \textit{Constraint Follow-Up} achieve high scores of 92.27\% and 92.36\%, respectively, while \textit{Dimension Rejection} improves from 42.37\% to 93.31\%. This confirms that the hybrid reward function, specifically through negative constraints on hallucination, successfully aligns the model to function as a rigorous agent that prioritizes semantic correctness over mere syntax generation.

\subsection{Ablation Study}
To quantify the contribution of each component in our hybrid reward mechanism, we conducted an ablation study by systematically removing specific auxiliary reward terms (Table~\ref{tab:ablation}). It is important to note that the primary execution reward $M_{\text{exec}}$ and the semantic similarity reward for natural language tasks were maintained across all settings, as they serve as the fundamental ground-truth signals for task correctness. Therefore, our analysis focuses on the impact of removing the supplementary constraints.

The results indicate that Grammatical Validity ($R_{\text{gram}}$) and Format Constraint ($R_{\text{fmt}}$) serve as foundational guardrails. Removing $R_{\text{gram}}$ causes the most significant performance degradation (-3.78\%), as the model loses the signal to generate executable SQL, leading to immediate failures in standard tasks. Similarly, the absence of $R_{\text{fmt}}$ (-3.39\%) disrupts the separation between reasoning traces and final answers, causing parsing failures particularly in complex \textit{Multi-Step Reasoning}.

In terms of optimization, Structure Similarity ($R_{\text{struct}}$) proves critical for complex SQL generation. Its removal leads to a sharp decline in \textit{Multi-Step Reasoning} tasks (from 81.69\% to 75.24\%) and \textit{Degenerate Dimension}, as the model lacks dense feedback on query structure, though it exhibits negligible impact on semantic tasks such as rejection. Finally, the Overlong Reward Shaping ($R_{\text{len}}$) functions as a stabilizer. While demonstrating the smallest average decrease (-1.82\%), it remains vital for the \textit{Reflection} task (decreasing to 90.63\%), preventing the model from entering verbose reasoning loops and ensuring efficient self-correction.

\begin{table*}[t]
\centering
\small
\caption{
Ablation study of different reward components.
Dim. Deg. denotes \textit{Degenerate Dimension}, and Dim. Rej. denotes \textit{Dimension Rejection}.
Avg. denotes the average performance across all tasks, and $\Delta$ represents the performance drop relative to the full Qwen3-GRPO model.
}
\label{tab:ablation}
\resizebox{\textwidth}{!}{
\begin{tabular}{lcccccccccc}
\toprule
\multirow{2}{*}{\textbf{Method Variant}} & \multicolumn{4}{c}{\textbf{SQL Generation Tasks}} & \multicolumn{4}{c}{\textbf{Boundary Interaction Tasks}} & \multirow{2}{*}{\textbf{Avg.}} & \multirow{2}{*}{\textbf{$\Delta$}} \\
\cmidrule(lr){2-5} \cmidrule(lr){6-9}
 & \textbf{Dim. Deg.} & \textbf{Reflection} & \textbf{Std. SQL} & \textbf{Multi-Step} & \textbf{Ambiguity} & \textbf{Dim. Rej.} & \textbf{Metric Rej.} & \textbf{Follow-Up} & & \\
\midrule
\textbf{BAR-SQL (Full)} & \textbf{93.22\%} & \textbf{93.75\%} & \textbf{90.75\%} & \textbf{81.69\%} & \textbf{94.51\%} & \textbf{93.31\%} & \textbf{92.27\%} & \textbf{92.36\%} & \textbf{91.48\%} & - \\
\midrule
w/o Grammar ($R_{\text{gram}}$) & 87.35\% & 92.71\% & 86.54\% & 80.21\% & 86.34\% & 87.37\% & 90.71\% & 90.40\% & 87.70\% & -3.78\% \\
w/o Format ($R_{\text{fmt}}$) & 90.01\% & 92.45\% & 87.69\% & 76.79\% & 89.43\% & 88.12\% & 89.15\% & 91.12\% & 88.10\% & -3.39\% \\
w/o Struct. Rule ($R_{\text{struct}}$) & 89.53\% & 91.63\% & 88.57\% & 75.24\% & 93.54\% & 93.31\% & 89.94\% & 91.12\% & 89.11\% & -2.37\% \\
w/o Overlong ($R_{\text{len}}$) & 92.88\% & 90.63\% & 89.69\% & 80.07\% & 92.08\% & 90.20\% & 90.71\% & 91.05\% & 89.66\% & -1.82\% \\
\bottomrule
\end{tabular}
}
\end{table*}

\subsection{Cognitive Paradigm Shift and Convergence in Reasoning}

Qualitative analysis of the training trajectory reveals that the performance gains afforded by GRPO extend beyond mere syntax optimization; they fundamentally reshape the model's latent reasoning paradigm. We observe a distinct evolution in the generated \texttt{<think>} traces, transitioning from rigid, mechanical rule-following to efficient, heuristic-driven expert intuition. This cognitive paradigm shift can be categorized into three distinct phases:

\textbf{1. From Linear Scanning to Cognitive Compression.} 
In the early training stages, the model exhibits ``risk-averse beginner'' behavior, particularly in \textit{Metric Rejection} tasks. To confirm the absence of a metric, early policies employ high-cost explicit reasoning, sequentially scanning every schema table and verbally confirming the absence of fields individually (e.g., ``Checked Table A, not found; Checked Table B, not found...''). As training converges, we observe a phenomenon of cognitive compression. The model learns to aggregate verification steps, adopting a set-based verification approach (e.g., ``Scanned the complete schema set; metric not found''). This shift substantially reduces inference latency while maintaining high certainty in refusal decisions.

\textbf{2. Suppression of Hallucination via Reward Dynamics.} 
A critical exploration spike is observed during the intermediate phase. Driven by the initial reward signal to generate SQL, the model transiently attempts to ``force'' answers for unanswerable queries by fabricating logic or misinterpreting semantic gaps (e.g., treating a customer ID as a business opportunity metric). However, under the adversarial penalty of the reward function for incorrect SQL generation, the policy rapidly self-corrects. The final paradigm shifts from ``attempting to solve everything'' to ``identifying the boundaries of solvability,'' thereby establishing a robust internal guardrail that suppresses the generation of invalid SQL with near-zero variance.

\textbf{3. Evolution from Over-Sensitivity to Intent-First Analysis.} 
In \textit{Ambiguity Clarification} task, early models display low tolerance for linguistic noise, frequently attempting to clarify irrelevant colloquialisms or minor obscurities. Through GRPO alignment, the model evolves into an active analyst. The converged model demonstrates an \textit{Intent-First} paradigm: it automatically filters input noise and focuses clarification solely on critical blockers, such as missing time ranges or undefined granularity. Similarly, for complex \textit{Multi-Step Reasoning}, the model transitions from ``drafting code'' within the thought process to ``high-level structural planning,'' separating logical topology from syntactic implementation.

To visualize this transformation, we summarize the structural evolution of the model's reasoning process in Table~\ref{table:paradigm_shift}.

\begin{table}[!t]
\caption{Comparison of thinking paradigms across training stages. The converged model exhibits structured, efficient, and boundary-aware reasoning compared to the linear and error-prone early stages.}
\centering
\small
\begin{tabular}{@{}p{0.95\columnwidth}@{}}
\toprule
\textbf{Phase I: Mechanical \& Verbose (Early Stage)} \\[2pt]
\textit{Pattern:} \texttt{[Entity Ext.]} $\rightarrow$ \texttt{[Table A Check]} $\rightarrow$ \dots $\rightarrow$ \texttt{[Draft Code]} $\rightarrow$ \texttt{[Decision]} \\[2pt]
\textit{Behavior:} Linearly scans all tables; interprets colloquial noise literally; drafts SQL syntax within reasoning traces. \\
\midrule
\textbf{Phase II: Exploration \& Hallucination (Intermediate Stage)} \\[2pt]
\textit{Pattern:} \texttt{[Entity Ext.]} $\rightarrow$ \texttt{[Forced Mapping]} $\rightarrow$ \texttt{[Logic Fabrication]} $\rightarrow$ \texttt{[Generate SQL]} \\[2pt]
\textit{Behavior:} Prioritizes action over correctness; attempts to bridge semantic gaps with fabricated logic to maximize potential rewards. \\
\midrule
\textbf{Phase III: Heuristic \& Structural (Converged Stage)} \\[2pt]
\textit{Pattern:} \texttt{[Intent Recog.]} $\rightarrow$ \texttt{[Constraint Check]} $\rightarrow$ \texttt{[Set Validation]} $\rightarrow$ \texttt{[Pattern Activation]} \\[2pt]
\textit{Behavior:} Logic compression (Set-based scanning); strictly adheres to business constraints; internalizes complex SQL structures (CTEs) as muscle memory. \\
\bottomrule
\end{tabular}
\label{table:paradigm_shift}
\end{table}

\section{Related Work}

\subsection{Evolution and Reliability in NL2SQL}

Natural Language to SQL (NL2SQL) serves as a critical bridge between database systems and natural language processing. Large language models have achieved remarkable performance on benchmarks such as Spider~\cite{spider1-yu2018spider} and BIRD~\cite{BIRD-li2023can}. However, as the field progresses toward more realistic and complex real-world environments, exemplified by Spider 2.0~\cite{spider2-lei2024spider}, a key challenge remains: when confronted with unanswerable queries, mainstream models tend to generate seemingly plausible yet incorrect SQL rather than choosing to abstain~\cite{drspider-chang2023dr}. To address this problem, several recent frameworks introduce external mechanisms. For instance, RTS~\cite{RTS-chen2025reliable} exploits conformal prediction to detect uncertainty, while systems such as ORANGE~\cite{orange-jiao2025orange} rely on retrieval to verify whether a query is consistent with domain knowledge. These approaches establish abstention capabilities but largely depend on the capacity of external discriminative modules or heuristic post-processing. In contrast, our work explores how to train models to intrinsically recognize boundary conditions, with the objective of enabling them to directly refuse or request clarification during the generation process, thereby providing a simplified alternative that obviates external orchestration.

\subsection{Composite Reasoning and Boundary-Aware Data Synthesis}

The reasoning capability of LLMs benefits substantially from diverse training distributions. Conventional data augmentation primarily focuses on valid (Question, SQL) pairs as positive supervision~\cite{data-jia2016data, grappa-yu2020grappa}, emphasizing alignment-oriented training while largely overlooking the instructional value of negative constraints and boundary cases. Chain-of-Thought methods and their variants~\cite{cot-wei2022chain, Rcot-xue2023rcot, sql2nl-hu2023importance, yoro-kobayashi2025you, Opensearch-sql-xie2025opensearch} expand training data through forward or backward generation strategies, enhancing reasoning ability under standard scenarios but remaining primarily confined to positive examples. 

Recent work has begun to recognize the value of modeling \textit{error boundaries}. Some studies distinguish executable from illegal SQL via execution feedback~\cite{PRACTIQ-dong-etal-2025-practiq, lever-ni2023lever}, or address unanswerable questions in dialogue~\cite{trustsql-lee2024trustsql}. Retry SQL~\cite{retry-sql-2025} further synthesizes retry-style chains of thought on BIRD by leveraging execution feedback to iteratively repair SQL, and reports systematic comparisons against proprietary LLMs. While this line of work highlights the benefit of exposing models to richer failure patterns, it continues to focus primarily on improving positive SQL success rates and remains constrained by the performance gap relative to closed-source systems. 

Building upon the above research, we propose a novel data synthesis pipeline that strictly anchors generated reasoning trajectories to schema definitions and business rules, while systematically constructing evolutionary data that encompasses both complex execution paths and boundary-testing scenarios (clarification, rejection, and follow-up). This enables the model to internalize boundary awareness during training, unifies the learning paradigms for generative and discriminative tasks, improves reasoning performance on complex queries, and effectively enhances transferability across heterogeneous domains.

\subsection{Reinforcement Learning Paradigms and Reward Engineering}

Current NL2SQL research predominantly follows two paradigms. Non-training approaches exploit sophisticated prompt engineering, Chain-of-Thought reasoning, and multi-agent collaboration to elicit the capabilities of large language models~\cite{dailsql-gao2023text, mac-sql-wang2025mac, chase-sql-pourreza2024ch}. In contrast, fine-tuning approaches demonstrate that while SFT alone may lead to memorization tendencies~\cite{excot-zhai2025excot}, combining it with Reinforcement Learning (RL) can enhance generalization~\cite{Sql-r1-ma2025sql, Reasoning-sql-pourreza2025reasoning, hes-sql-qiu2025hes}. Among RL algorithms, Proximal Policy Optimization (PPO)~\cite{ppo-schulman2017proximal} remains widely adopted, whereas GRPO~\cite{grpo-shao2024deepseekmath} eliminates the value network via group-relative advantage estimation, thereby reducing memory overhead and improving training stability in resource-constrained environments, offering an attractive alternative.

The design of reward signals likewise determines the upper bound of RL-based model capabilities. Early methods relied on exact matching of execution results~\cite{seq2sql-zhong2017seq2sql}, which led to the problem of sparse rewards~\cite{memory-liang2018memory}. Recent work has introduced finer-grained designs: partial rewards provide intermediate feedback along the reasoning trajectory~\cite{Reasoning-sql-pourreza2025reasoning}, and Abstract Syntax Tree (AST) analysis enables structural evaluation~\cite{learnat-liao2025learnat}. However, existing reward mechanisms primarily optimize generation accuracy and often neglect the reliability of abstention. We introduce a novel task-condition-aware hybrid reward mechanism aimed at balancing the dual objectives of SQL execution and conversational refusal, enabling a single policy to effectively adapt to diverse query types in industrial-scale scenarios.

\section{Conclusion}
Commercial deployment of NL2SQL is currently hindered by the dual necessity of high execution accuracy and stringent reliability. While existing LLMs demonstrate promise, they struggle to meet the zero-tolerance standards of enterprise Business Intelligence (BI), frequently failing to balance complex reasoning with awareness of knowledge boundaries. To bridge this gap, we propose BAR-SQL, a unified generative framework that integrates \textit{Seed-Mutation} data synthesis, \textit{Knowledge-Grounded Reasoning Synthesis} (KGRS), and a GRPO-based Reinforcement Learning strategy. Extensive experiments demonstrate that our approach achieves 91.48\% accuracy, substantially outperforming proprietary state-of-the-art models by simultaneously mastering complex SQL generation and precise abstention.

Beyond empirical gains, our work validates the efficacy of a unified generative paradigm for enterprise NL2SQL. We demonstrate that by synthesizing high-fidelity boundary constraints and enforcing them through Reinforcement Learning, models can internalize the distinction between answerable and unanswerable queries. This transforms the NL2SQL agent from a passive code translator into a proactive collaborator that rigorously adheres to business logic without relying on external discrimination modules.

A remaining limitation is our reliance on outcome-based supervision, which leaves intermediate reasoning steps unverified. Inspired by recent advances in reasoning supervision~\cite{shao2025deepseekmath}, future work will explore integrating a Reasoning-Aware Verification Reward Model into the NL2SQL context. Additionally, while our design prioritizes efficiency under practical deployment constraints, exploring larger-capacity models may yield incremental performance improvements.


\bibliographystyle{ACM-Reference-Format}
\bibliography{NL2SQL_reference}

@article{zhang2024sqlfuse,
  title={Sqlfuse: Enhancing text-to-sql performance through comprehensive llm synergy},
  author={Zhang, Tingkai and Chen, Chaoyu and Liao, Cong and Wang, Jun and Zhao, Xudong and Yu, Hang and Wang, Jianchao and Li, Jianguo and Shi, Wenhui},
  journal={arXiv preprint arXiv:2407.14568},
  year={2024}
}

@article{shao2025deepseekmath,
  title={DeepSeekMath-V2: Towards Self-Verifiable Mathematical Reasoning},
  author={Shao, Zhihong and Luo, Yuxiang and Lu, Chengda and Ren, ZZ and Hu, Jiewen and Ye, Tian and Gou, Zhibin and Ma, Shirong and Zhang, Xiaokang},
  journal={arXiv preprint arXiv:2511.22570},
  year={2025}
}

@article{ding2025ambisql,
  title={AmbiSQL: Interactive Ambiguity Detection and Resolution for Text-to-SQL},
  author={Ding, Zhongjun and Lin, Yin and Zeng, Tianjing},
  journal={arXiv preprint arXiv:2508.15276},
  year={2025}
}

@online{stonebraker2025dbos,
  author       = {Mike Stonebraker and Andy Pavlo},
  title        = {Data 2025: The Year in Review with Mike Stonebraker \& Andy Pavlo},
  year         = {2025},
  month        = {Dec.},
  url          = {https://www.dbos.dev/webcast-2025-in-review-with-mike-stonebraker-and-andy-pavlo},
  organization = {DBOS, Inc.},
  note         = {Official webcast page with description of the 2025 database trends review},
}

@article{wang2025fdabench,
  title={FDABench: A Benchmark for Data Agents on Analytical Queries over Heterogeneous Data},
  author={Wang, Ziting and Zhang, Shize and Yuan, Haitao and Zhu, Jinwei and Li, Shifu and Dong, Wei and Cong, Gao},
  journal={arXiv preprint arXiv:2509.02473},
  year={2025}
}

@article{liu2025survey,
  title={A survey of text-to-sql in the era of llms: Where are we, and where are we going?},
  author={Liu, Xinyu and Shen, Shuyu and Li, Boyan and Ma, Peixian and Jiang, Runzhi and Zhang, Yuxin and Fan, Ju and Li, Guoliang and Tang, Nan and Luo, Yuyu},
  journal={IEEE Transactions on Knowledge and Data Engineering},
  year={2025},
  publisher={IEEE}
}

@article{chen2025reliable,
  title={Reliable Text-to-SQL with Adaptive Abstention},
  author={Chen, Kaiwen and Chen, Yueting and Koudas, Nick and Yu, Xiaohui},
  journal={Proceedings of the ACM on Management of Data},
  volume={3},
  number={1},
  pages={1--30},
  year={2025},
  publisher={ACM New York, NY, USA}
}

@article{hes-sql-qiu2025hes,
  title={HES-SQL: Hybrid Reasoning for Efficient Text-to-SQL with Structural Skeleton Guidance},
  author={Qiu, Suming and Li, Jing and Zhou, Zhicheng and Huang, Junjie and Qiu, Linyuan and Sun, Zhijie},
  journal={arXiv preprint arXiv:2510.08896},
  year={2025}
}

@article{Sql-r1-ma2025sql,
  title={Sql-r1: Training natural language to sql reasoning model by reinforcement learning},
  author={Ma, Peixian and Zhuang, Xialie and Xu, Chengjin and Jiang, Xuhui and Chen, Ran and Guo, Jian},
  journal={arXiv preprint arXiv:2504.08600},
  year={2025}
}

@inproceedings{zhong2020grounded,
  title={Grounded Adaptation for Zero-shot Executable Semantic Parsing},
  author={Zhong, Victor and Lewis, Mike and Wang, Sida I and Zettlemoyer, Luke},
  booktitle={Proceedings of the 2020 Conference on Empirical Methods in Natural Language Processing (EMNLP)},
  pages={6869--6882},
  year={2020}
}

@article{chase-sql-pourreza2024ch,
  title={Chase-sql: Multi-path reasoning and preference optimized candidate selection in text-to-sql},
  author={Pourreza, Mohammadreza and Li, Hailong and Sun, Ruoxi and Chung, Yeounoh and Talaei, Shayan and Kakkar, Gaurav Tarlok and Gan, Yu and Saberi, Amin and Ozcan, Fatma and Arik, Sercan O},
  journal={arXiv preprint arXiv:2410.01943},
  year={2024}
}

@article{hwang2019comprehensive,
  title={A comprehensive exploration on wikisql with table-aware word contextualization},
  author={Hwang, Wonseok and Yim, Jinyeong and Park, Seunghyun and Seo, Minjoon},
  journal={arXiv preprint arXiv:1902.01069},
  year={2019}
}

@article{sadga-cai2021sadga,
  title={Sadga: Structure-aware dual graph aggregation network for text-to-sql},
  author={Cai, Ruichu and Yuan, Jinjie and Xu, Boyan and Hao, Zhifeng},
  journal={Advances in Neural Information Processing Systems},
  volume={34},
  pages={7664--7676},
  year={2021}
}

@article{dailsql-gao2023text,
  title={Text-to-sql empowered by large language models: A benchmark evaluation},
  author={Gao, Dawei and Wang, Haibin and Li, Yaliang and Sun, Xiuyu and Qian, Yichen and Ding, Bolin and Zhou, Jingren},
  journal={arXiv preprint arXiv:2308.15363},
  year={2023}
}

@article{cot-wei2022chain,
  title={Chain-of-thought prompting elicits reasoning in large language models},
  author={Wei, Jason and Wang, Xuezhi and Schuurmans, Dale and Bosma, Maarten and Xia, Fei and Chi, Ed and Le, Quoc V and Zhou, Denny and others},
  journal={Advances in neural information processing systems},
  volume={35},
  pages={24824--24837},
  year={2022}
}

@article{Rcot-xue2023rcot,
  title={Rcot: Detecting and rectifying factual inconsistency in reasoning by reversing chain-of-thought},
  author={Xue, Tianci and Wang, Ziqi and Wang, Zhenhailong and Han, Chi and Yu, Pengfei and Ji, Heng},
  journal={arXiv preprint arXiv:2305.11499},
  year={2023}
}

@article{BIRD-li2023can,
  title={Can llm already serve as a database interface? a big bench for large-scale database grounded text-to-sqls},
  author={Li, Jinyang and Hui, Binyuan and Qu, Ge and Yang, Jiaxi and Li, Binhua and Li, Bowen and Wang, Bailin and Qin, Bowen and Geng, Ruiying and Huo, Nan and others},
  journal={Advances in Neural Information Processing Systems},
  volume={36},
  pages={42330--42357},
  year={2023}
}

@article{trustsql-lee2024trustsql,
  title={Trustsql: A reliability benchmark for text-to-sql models with diverse unanswerable questions},
  author={Lee, Gyubok and Chay, Woosog and Cho, Seonhee and Choi, Edward},
  journal={CoRR},
  year={2024}
}

@article{spider1-yu2018spider,
  title={Spider: A large-scale human-labeled dataset for complex and cross-domain semantic parsing and text-to-sql task},
  author={Yu, Tao and Zhang, Rui and Yang, Kai and Yasunaga, Michihiro and Wang, Dongxu and Li, Zifan and Ma, James and Li, Irene and Yao, Qingning and Roman, Shanelle and others},
  journal={arXiv preprint arXiv:1809.08887},
  year={2018}
}

@inproceedings{spider2-lei2024spider,
  title={Spider 2.0: Evaluating Language Models on Real-World Enterprise Text-to-SQL Workflows},
  author={Lei, Fangyu and Chen, Jixuan and Ye, Yuxiao and Cao, Ruisheng and Shin, Dongchan and SU, Hongjin and SUO, ZHAOQING and Gao, Hongcheng and Hu, Wenjing and Yin, Pengcheng and others},
  booktitle={The Thirteenth International Conference on Learning Representations},
}

@article{RTS-chen2025reliable,
  title={Reliable Text-to-SQL with Adaptive Abstention},
  author={Chen, Kaiwen and Chen, Yueting and Koudas, Nick and Yu, Xiaohui},
  journal={Proceedings of the ACM on Management of Data},
  volume={3},
  number={1},
  pages={1--30},
  year={2025},
  publisher={ACM New York, NY, USA}
}

@article{orange-jiao2025orange,
  title={ORANGE: An Online Reflection ANd GEneration framework with Domain Knowledge for Text-to-SQL},
  author={Jiao, Yiwen and Ren, Tonghui and Gao, Yuche and He, Zhenying and Jing, Yinan and Zhang, Kai and Wang, X Sean},
  journal={arXiv preprint arXiv:2511.00985},
  year={2025}
}

@inproceedings{mac-sql-wang2025mac,
  title={Mac-sql: A multi-agent collaborative framework for text-to-sql},
  author={Wang, Bing and Ren, Changyu and Yang, Jian and Liang, Xinnian and Bai, Jiaqi and Chai, Linzheng and Yan, Zhao and Zhang, Qian-Wen and Yin, Di and Sun, Xing and others},
  booktitle={Proceedings of the 31st International Conference on Computational Linguistics},
  pages={540--557},
  year={2025}
}

@article{Reasoning-sql-pourreza2025reasoning,
  title={Reasoning-sql: Reinforcement learning with sql tailored partial rewards for reasoning-enhanced text-to-sql},
  author={Pourreza, Mohammadreza and Talaei, Shayan and Sun, Ruoxi and Wan, Xingchen and Li, Hailong and Mirhoseini, Azalia and Saberi, Amin and Arik, Sercan and others},
  journal={arXiv preprint arXiv:2503.23157},
  year={2025}
}

@article{grpo-shao2024deepseekmath,
  title={Deepseekmath: Pushing the limits of mathematical reasoning in open language models},
  author={Shao, Zhihong and Wang, Peiyi and Zhu, Qihao and Xu, Runxin and Song, Junxiao and Bi, Xiao and Zhang, Haowei and Zhang, Mingchuan and Li, YK and Wu, Yang and others},
  journal={arXiv preprint arXiv:2402.03300},
  year={2024}
}

@article{Opensearch-sql-xie2025opensearch,
  title={Opensearch-sql: Enhancing text-to-sql with dynamic few-shot and consistency alignment},
  author={Xie, Xiangjin and Xu, Guangwei and Zhao, Lingyan and Guo, Ruijie},
  journal={Proceedings of the ACM on Management of Data},
  volume={3},
  number={3},
  pages={1--24},
  year={2025},
  publisher={ACM New York, NY, USA}
}

@inproceedings{sql2nl-hu2023importance,
  title={Importance of synthesizing high-quality data for text-to-SQL parsing},
  author={Hu, Yiqun and Zhao, Yiyun and Jiang, Jiarong and Lan, Wuwei and Zhu, Henghui and Chauhan, Anuj and Li, Alexander Hanbo and Pan, Lin and Wang, Jun and Hang, Chung-Wei and others},
  booktitle={Findings of the Association for Computational Linguistics: ACL 2023},
  pages={1327--1343},
  year={2023}
}

@inproceedings{yoro-kobayashi2025you,
  title={You only read once (YORO): Learning to internalize database knowledge for text-to-SQL},
  author={Kobayashi, Hideo and Lan, Wuwei and Shi, Peng and Chang, Shuaichen and Guo, Jiang and Zhu, Henghui and Wang, Zhiguo and Ng, Patrick},
  booktitle={Proceedings of the 2025 Conference of the Nations of the Americas Chapter of the Association for Computational Linguistics: Human Language Technologies (Volume 1: Long Papers)},
  pages={1889--1901},
  year={2025}
}

@inproceedings{PRACTIQ-dong-etal-2025-practiq,
    title = "{PRACTIQ}: A Practical Conversational Text-to-{SQL} dataset with Ambiguous and Unanswerable Queries",
    author = "Dong, Mingwen  and
      Ashok Kumar, Nischal  and
      Hu, Yiqun  and
      Chauhan, Anuj  and
      Hang, Chung-Wei  and
      Chang, Shuaichen  and
      Pan, Lin  and
      Lan, Wuwei  and
      Zhu, Henghui  and
      Jiang, Jiarong  and
      Ng, Patrick  and
      Wang, Zhiguo",
    editor = "Chiruzzo, Luis  and
      Ritter, Alan  and
      Wang, Lu",
    booktitle = "Proceedings of the 2025 Conference of the Nations of the Americas Chapter of the Association for Computational Linguistics: Human Language Technologies (Volume 1: Long Papers)",
    month = apr,
    year = "2025",
    address = "Albuquerque, New Mexico",
    publisher = "Association for Computational Linguistics",
    url = "https://aclanthology.org/2025.naacl-long.13/",
    doi = "10.18653/v1/2025.naacl-long.13",
    pages = "255--273",
    ISBN = "979-8-89176-189-6"
}

@inproceedings{lever-ni2023lever,
  title={Lever: Learning to verify language-to-code generation with execution},
  author={Ni, Ansong and Iyer, Srini and Radev, Dragomir and Stoyanov, Veselin and Yih, Wen-tau and Wang, Sida and Lin, Xi Victoria},
  booktitle={International Conference on Machine Learning},
  pages={26106--26128},
  year={2023},
  organization={PMLR}
}

@article{yu2025dapo,
  title={Dapo: An open-source llm reinforcement learning system at scale},
  author={Yu, Qiying and Zhang, Zheng and Zhu, Ruofei and Yuan, Yufeng and Zuo, Xiaochen and Yue, Yu and Dai, Weinan and Fan, Tiantian and Liu, Gaohong and Liu, Lingjun and others},
  journal={arXiv preprint arXiv:2503.14476},
  year={2025}
}

@misc{qwen3technicalreport,
      title={Qwen3 Technical Report}, 
      author={Qwen Team},
      year={2025},
      eprint={2505.09388},
      archivePrefix={arXiv},
      primaryClass={cs.CL},
      url={https://arxiv.org/abs/2505.09388}, 
}

@article{guo2025deepseek,
  title={Deepseek-r1 incentivizes reasoning in llms through reinforcement learning},
  author={Guo, Daya and Yang, Dejian and Zhang, Haowei and Song, Junxiao and Wang, Peiyi and Zhu, Qihao and Xu, Runxin and Zhang, Ruoyu and Ma, Shirong and Bi, Xiao and others},
  journal={Nature},
  volume={645},
  number={8081},
  pages={633--638},
  year={2025},
  publisher={Nature Publishing Group UK London}
}

@article{drspider-chang2023dr,
  title={Dr. spider: A diagnostic evaluation benchmark towards text-to-sql robustness},
  author={Chang, Shuaichen and Wang, Jun and Dong, Mingwen and Pan, Lin and Zhu, Henghui and Li, Alexander Hanbo and Lan, Wuwei and Zhang, Sheng and Jiang, Jiarong and Lilien, Joseph and others},
  journal={arXiv preprint arXiv:2301.08881},
  year={2023}
}

@article{data-jia2016data,
  title={Data recombination for neural semantic parsing},
  author={Jia, Robin and Liang, Percy},
  journal={arXiv preprint arXiv:1606.03622},
  year={2016}
}

@article{grappa-yu2020grappa,
  title={Grappa: Grammar-augmented pre-training for table semantic parsing},
  author={Yu, Tao and Wu, Chien-Sheng and Lin, Xi Victoria and Wang, Bailin and Tan, Yi Chern and Yang, Xinyi and Radev, Dragomir and Socher, Richard and Xiong, Caiming},
  journal={arXiv preprint arXiv:2009.13845},
  year={2020}
}

@article{ppo-schulman2017proximal,
  title={Proximal policy optimization algorithms},
  author={Schulman, John and Wolski, Filip and Dhariwal, Prafulla and Radford, Alec and Klimov, Oleg},
  journal={arXiv preprint arXiv:1707.06347},
  year={2017}
}

@article{learnat-liao2025learnat,
  title={LearNAT: Learning NL2SQL with AST-guided Task Decomposition for Large Language Models},
  author={Liao, Weibin and Gao, Xin and Jia, Tianyu and Qiu, Rihong and Zhu, Yifan and Lin, Yang and Chu, Xu and Zhao, Junfeng and Wang, Yasha},
  journal={arXiv preprint arXiv:2504.02327},
  year={2025}
}

@article{excot-zhai2025excot,
  title={ExCoT: Optimizing reasoning for text-to-SQL with execution feedback},
  author={Zhai, Bohan and Xu, Canwen and He, Yuxiong and Yao, Zhewei},
  journal={arXiv preprint arXiv:2503.19988},
  year={2025}
}

@article{seq2sql-zhong2017seq2sql,
  title={Seq2sql: Generating structured queries from natural language using reinforcement learning},
  author={Zhong, Victor and Xiong, Caiming and Socher, Richard},
  journal={arXiv preprint arXiv:1709.00103},
  year={2017}
}

@article{memory-liang2018memory,
  title={Memory augmented policy optimization for program synthesis and semantic parsing},
  author={Liang, Chen and Norouzi, Mohammad and Berant, Jonathan and Le, Quoc V and Lao, Ni},
  journal={Advances in Neural Information Processing Systems},
  volume={31},
  year={2018}
}

@article{retry-sql-2025,
  title={RetrySQL: text-to-SQL training with retry data for self-correcting query generation},
  author={Belluzzo, Riccardo and Baran, Joanna and Olszewski, Pawe{\'L} and others},
  journal={arXiv preprint arXiv:2507.02529},
  year={2025}
}

\newpage
\newpage
\appendix

\section{Application in Business Intelligence}
\label{app:bi_application}

A query-type classifier first filters user questions, forwarding only BI/SQL-related queries to the NL2SQL stack while rejecting out-of-scope requests.
For valid queries, a metric-keyed retrieval module injects schema, metric, domain, and data knowledge into the prompt, and the model generates either executable SQL, clarification questions, or knowledge-based refusals with optional reflection for error correction.
This design decouples business knowledge from the model: practitioners can flexibly customize query logic by updating fields and metric definitions, achieving stable query behavior while maintaining high extensibility as schemas and indicators evolve.


\begin{figure*}[t]
\centering
\includegraphics[width=1.0\textwidth]{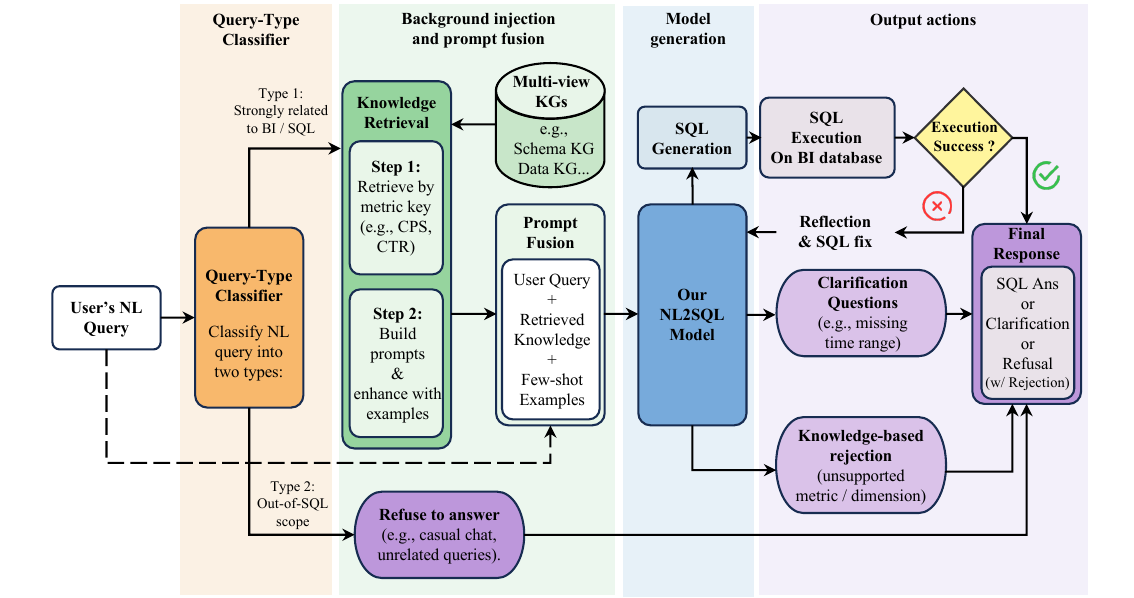}
\caption{Application pipeline of our NL2SQL system.}
\label{fig:ApplicationPipeline}
\end{figure*}



\section{Data System Construction}
\label{app:data_sys}

\subsection{Natural-Language Question Generation with Emergent Augmentation}

During seed generation, slot-filled template instances are transformed into natural business questions by an LLM guided by the Knowledge Context. Rather than enforcing a fixed distribution of query types, we employ \textit{emergent augmentation}: the LLM naturally introduces analytical variations (e.g., Top-K ranking, extremum selection) that fit the business scenario, constrained by the available schema, metrics, and domain rules. For instance, \texttt{[Year 2025, Li Auto L6, Leads, By Month]} may yield direct aggregation (\textit{"What is the monthly lead count..."}), extremum queries (\textit{"Which month had the highest..."}), or ranking (\textit{"Show me the top-3 months..."}). The prompt template is provided in Figure~\ref{fig:nl-question-generation-prompt}.

\subsection{Examples of Synthesized Data}
Table~\ref{tab:synthesis_examples} illustrates representative examples from our synthesized corpus, spanning from standard SQL generation to complex reasoning and interactive boundary awareness. Crucially, the table highlights the role of the KGRS paradigm: for every task type, the synthesized \texttt{<think>} trace explicitly cites schema definitions or domain rules to justify the final output. This demonstrates how our pipeline enforces evidence-based reasoning across both positive code generation and negative constraint-handling scenarios.

\begin{table*}[ht]
\centering
\small
\renewcommand{\arraystretch}{1.2}
\caption{Examples of Synthesized Training Data across Different Tasks. The \texttt{<think>} traces are synthesized via the \textbf{KGRS paradigm}, explicitly grounding the reasoning in the provided knowledge context.}
\label{tab:synthesis_examples}
\begin{tabular}{p{0.13\linewidth} p{0.24\linewidth} p{0.57\linewidth}}
\toprule
\textbf{Task Type} & \textbf{Example Question} & \textbf{Synthesized Output (CoT \& Response)} \\
\midrule
\textbf{Standard SQL} & Target test drive volume for the first week of 2023. & \texttt{<think>} [KGRS] Analyze time ('202301') and metric ('target\_cnt'). Construct query with filters... \texttt{</think>} \newline \texttt{<answer>} SELECT SUM(target\_cnt) ... WHERE t2.year\_natural\_week\_id = '202301' \texttt{</answer>} \\
\hline
\textbf{Multi-Step \newline Reasoning} & Leads share increase in Jiangsu (2024 vs 2023). & \texttt{<think>} [KGRS] Use CTEs for YoY analysis. CTE1: 2023 stats; CTE2: 2024 stats. Compute difference... \texttt{</think>} \newline \texttt{<answer>} WITH year\_channel\_stats AS (...) SELECT ... \texttt{</answer>} \\
\hline
\textbf{Reflection} & [Input: Error SQL + Error Msg: 'Column t4.series\_name cannot be resolved'] & \texttt{<think>} [KGRS] Error indicates identifier resolution failure. The subquery scope does not export 'series\_name'. I must add it to the projection... \texttt{</think>} \newline \texttt{<answer>} SELECT ... FROM (SELECT series\_name ...) ... \texttt{</answer>} \\
\hline
\textbf{Degenerate \newline Dimension} & Target volume by province. (Schema degraded to flat table) & \texttt{<think>} [KGRS] Schema check: `province\_name` is now in fact table `dwd\_sale\_target`. No join with `dim\_area` needed. \texttt{</think>} \newline \texttt{<answer>} SELECT province\_name, SUM(target\_cnt) ... GROUP BY province\_name \texttt{</answer>} \\
\hline
\textbf{Ambiguity \newline Clarification} & Indicators for provinces like Inner Mongolia recently? & \texttt{<think>} [KGRS] "Recently" is vague. "Indicators" is undefined. Cannot generate SQL. \texttt{</think>} \newline \texttt{<answer>} Could you specify the time range and the specific indicator? \texttt{</answer>} \\
\hline
\textbf{Dimension \newline Rejection} & Test drive volume for L9 series from 2022 to 2024. & \texttt{<think>} [KGRS] Time range valid. Schema check: No field represents 'L9 Series'. Cannot filter. \texttt{</think>} \newline \texttt{<answer>} Due to the lack of a defined field for the L9 series, this query is not supported. \texttt{</answer>} \\
\hline
\textbf{Metric Rejection} & High-intent leads targets for the last year. & \texttt{<think>} [KGRS] Check schema for 'High-intent Leads'. No such metric found in `dwd\_sale\_target`. \texttt{</think>} \newline \texttt{<answer>} Due to the lack of a defined metric, this query is not supported. \texttt{</answer>} \\
\hline
\textbf{Constraint \newline Follow-Up} & Show net new leads for this month by week. & \texttt{<think>} [KGRS] Metric 'Net New Leads' mandates a 'Car Series' constraint. Query lacks this info. \texttt{</think>} \newline \texttt{<answer>} Please specify the Car Series when querying Net New Leads. \texttt{</answer>} \\
\bottomrule
\end{tabular}
\end{table*}

\subsection{Prompts for Data Synthesis and Inference}
This part provides the core prompt templates used in our framework. First, we present the templates designed for CoT Synthesis, which instantiate the KGRS paradigm. These prompts are engineered to elicit structured, reproducible reasoning traces that are explicitly anchored to the provided knowledge context (schema, metrics, rules) to generate high-quality training data:

\begin{itemize}
    \item \textbf{Multi-Step Inference CoT Synthesis} (Fig.~\ref{fig:multi-step-cot-prompt}): Guides the decomposition of a complex analytical question and produces a detailed, first-person narrative explaining the stepwise logic and SQL design choices (e.g., CTEs, window functions).
    \item \textbf{Reflection CoT Synthesis} (Fig.~\ref{fig:reflection-cot-prompt}): Generates a debugging trace by analyzing an erroneous SQL query and its error message to synthesize a correction path.
    \item \textbf{Degenerate Dimension CoT Synthesis} (Fig.~\ref{fig:degenerate-dimension-cot-prompt}): Generates reasoning traces for adapting SQL queries to denormalized schemas where dimension attributes have been physically merged into fact tables, eliminating the need for JOIN operations while maintaining equivalent business logic.
    \item \textbf{Clarification CoT Synthesis} (Fig.~\ref{fig:clarification-cot-prompt}): Produces reasoning to identify ambiguity in a user question and formulates a concise clarification request based on the available schema.
    \item \textbf{Dimension Rejection CoT Synthesis} (Fig.~\ref{fig:dimension-rejection-cot-prompt}): Constructs a rationale for rejecting a query due to missing dimension fields in the schema.
    \item \textbf{Metric Rejection CoT Synthesis} (Fig.~\ref{fig:metric-rejection-prompt}): Constructs a rationale for rejecting a query due to an undefined metric.
    \item \textbf{Follow-up CoT Synthesis} (Fig.~\ref{fig:follow-up-cot-prompt}): Generates reasoning to identify a missing mandatory constraint and asks the user for the required parameter.
\end{itemize}

In addition to the synthesis templates above, we provide distinct templates for validation and deployment. \textbf{SQL Validation} (Fig.~\ref{fig:sql-validation-prompt}) guides a stepwise analysis to judge if a SQL query correctly answers a business question. Finally, \textbf{Model Inference} (Fig.~\ref{fig:inference-prompt}) serves as the unified template for deployment, integrating all knowledge sources in a few-shot format to guide the model's \texttt{<think>} and \texttt{<answer>} generation.

\begin{figure*}[!h]
    \centering
    \begin{promptbox}{Prompt for Natural-Language Question Generation with Emergent Augmentation}
\textbf{--- 0. Background Knowledge Context}\\
\textbf{\#\#\# Schema Knowledge (Schema KG)}\\
\texttt{\{schema\_info\}}\\
{\small\color{gray}\textit{Complete DDL definitions including fact tables, dimension tables, primary/foreign keys, and semantic field descriptions.}}\\

\textbf{\#\#\# Metric Knowledge (Metric KG)}\\
\texttt{\{metric\_definitions\}}\\
{\small\color{gray}\textit{Business metric definitions with aliases, mandatory constraints, and computation rules (e.g., "Net New Leads" requires \texttt{is\_net\_leads=1} and \texttt{count(distinct customer\_account\_id)}).}}\\

\textbf{\#\#\# Domain Knowledge (Domain KG)}\\
\texttt{\{domain\_rules\}}\\
{\small\color{gray}\textit{Business rules for time range interpretation, grouping defaults, and analytical conventions (e.g., "last 7 days", default channel granularity).}}\\

\textbf{\#\#\# Data Knowledge (Data KG)}\\
\texttt{\{entity\_mappings\}}\\
{\small\color{gray}\textit{Entity-to-field mappings such as channel hierarchies and geographic attributes.}}\\

\textbf{--- 1. Role and Task}\\
You are a business analyst fluent in domain terminology. Your task is to transform the given slot-filled structure into a natural, human-style business question that aligns with real-world BI scenarios.\\

\textbf{--- 2. Input: Slot-Filled Structure}\\
\texttt{\{template\_instance\}}\\
{\small\color{gray}\textit{Example: [Time: Year 2025, Brand: Li Auto L6, Metric: Leads, GroupBy: By Month]}}\\

\textbf{--- 3. Generation Guidelines}\\
\textbf{(1) Natural Expression}: Generate fluent questions as a business user would ask, not mechanical template translations.\\

\textbf{(2) Emergent Augmentation}: You may naturally introduce analytical variations that fit the business context, such as:\\
\ \ \ \ $\bullet$ \textit{Ranking queries}: "Show me the top-3 months..."\\
\ \ \ \ $\bullet$ \textit{Extremum selection}: "Which month had the highest..."\\
\ \ \ \ $\bullet$ \textit{Direct aggregation}: "What is the total/average..."\\
\ \ \ \ $\bullet$ \textit{Trend analysis}: "How did ... change over time..."\\

\textbf{(3) Knowledge Constraints}: All entities, metrics, and analytical operations must reference only elements defined in the provided Knowledge Context. Do not introduce undefined dimensions or metrics.\\

\textbf{(4) No Fixed Distribution}: Augmentation types should emerge naturally based on business relevance—do not force a predefined distribution.\\

\textbf{--- 4. Output Examples}\\
\textbf{Input}: [Time: Year 2025, Brand: Li Auto L6, Metric: Leads, GroupBy: By Month]\\

\textbf{Valid Outputs}:\\
\ \ \ \ $\bullet$ "What is the monthly lead count for Li Auto L6 in 2025?" \textit{(direct translation)}\\
\ \ \ \ $\bullet$ "In 2025, which month had the highest leads for Li Auto L6?" \textit{(extremum)}\\
\ \ \ \ $\bullet$ "Show me the top-3 months by lead count for Li Auto L6 in 2025." \textit{(ranking)}\\

\textbf{--- 5. Start Task}\\
Please generate a natural-language question based on the slot-filled structure above. Output only the question text without additional explanations.
    \end{promptbox}
    \caption{Prompt for Natural-Language Question Generation with Emergent Augmentation}
    \label{fig:nl-question-generation-prompt}
\end{figure*}

\begin{figure*}[!h]
    \centering
    \begin{promptbox}{Prompt for Multi-Step Inference COT Generation}
\textbf{--- 0. Background Knowledge}\\
Database Schema: \texttt{\{schema\_info\}}\\
Knowledge Graph and Business Logic: \texttt{\{knowledge\_info\}}\\
Domain Knowledge and Rules: \texttt{\{domain\_info\}}\\
Supplementary Knowledge: \texttt{\{data\_kg\}}\\

\textbf{--- 1. Role and Task}\\
You are a senior data analytics architect who excels at decomposing complex multi-layered business problems into clear analytical logic and can deeply explain the construction approach and implementation details of complex SQL queries. Your task is to generate a detailed, in-depth, and logically coherent chain-of-thought (COT) reasoning process based on the context, business analysis question, and validated SQL query provided. This reasoning process should fully demonstrate your entire reasoning journey from understanding the problem to designing the SQL structure and then to implementation details.\\

\textbf{--- 2. Core Instructions}\\
\textbf{(1) First-Person Narrative}: Your reasoning process should resemble a senior analyst's internal monologue when solving problems. Please narrate in first person ("I").\\

\textbf{(2) Natural Analysis Flow}: Your analysis should unfold naturally, starting from understanding the core requirements, complexity, and ultimate goals of the problem, then analyzing what data tables are needed and their associations, and finally explaining how filtering conditions, metric calculations, and grouping dimensions are determined based on business needs.\\

\textbf{(3) Deep Explanation of Complex Logic}: When SQL uses complex logic (such as CTEs, window functions, or multi-layer comparative calculations), you \textbf{must explain in depth why} you chose this technique and how it helps you construct the final result step by step. For example, when performing "above average" comparisons, you need to clearly explain how the "average value" baseline is calculated; when using CTEs, you should explain the purpose of each CTE and its advantages compared to alternative approaches.\\

\textbf{(4) Strictly Follow Narrative Style}: \textbf{Most importantly}, your final output \textbf{must strictly mimic} the narrative style below:\\
\textit{Alright, let me analyze this problem in depth. First,... Next,... Then,... Finally,...}\\

\textbf{--- 3. Quality Reference Example (Complex Multi-layer Calculation Scenario)}\\
Alright, let me analyze this problem in depth. The question requires finding, within the most recent half quarter for the "L Series" product line, which provinces have "central store" net new lead proportions higher than the national average proportion in second and third-tier cities, where the province's "central store" net new leads are at least 500. Finally, I need to output province name, central store net new lead count, provincial proportion, national average proportion, and proportion difference, sorted by proportion difference in descending order.\\

First, I need to understand the complexity of this problem. This is a multi-layered comparative analysis problem. It involves three levels of calculation: the first level calculates each province's central store net new lead count and total lead count; the second level calculates each province's proportion; the third level calculates the national average proportion as a comparison baseline.\\

Next, ...\\

For SQL structure design, ...\\

Finally, I need to calculate the proportion difference (provincial proportion - national average proportion) in the results, filter records with positive differences (i.e., provinces above the national average), and then sort by difference in descending order.\\

\textbf{--- 4. Start Task}\\
\textbf{Question}: \texttt{\{question\}}\\
\textbf{SQL Query}:\\
\texttt{`\hspace{0.05mm}`\hspace{0.05mm}`sql}\\
\texttt{\{sql\}}\\
\texttt{`\hspace{0.05mm}`\hspace{0.05mm}`}\\

\textbf{--- 5. Output Format}\\
Directly output the reasoning process text content without any other instructions or formatting. The reasoning process should be detailed and in-depth, fully demonstrating understanding and analytical capability of complex SQL logic.
    \end{promptbox}
    \caption{Prompt for Multi-Step Inference COT Generation}
    \label{fig:multi-step-cot-prompt}
\end{figure*}

\begin{figure*}[!h]
    \centering
    \begin{promptbox}{Prompt for Reflection COT Synthesis}
\textbf{--- 0. Background Knowledge}\\
Database Schema: \texttt{\{schema\_info\}}\\
Knowledge Graph: \texttt{\{knowledge\_info\}}\\
Domain Rules: \texttt{\{domain\_info\}}\\

\textbf{--- 1. Role and Task}\\
You are an experienced data analyst. Your task is to analyze an SQL execution error and perform reflective correction based on the given background information, user question, erroneous SQL, error message, and correct SQL.\\

\textbf{--- 2. Core Instructions}\\
Please simulate a step-by-step reflection and correction process, including:\\

\textbf{(1) Error Analysis}: Analyze the error message in detail and identify the specific type of error (e.g., syntax error, identifier error, runtime error, etc.).\\

\textbf{(2) Error Localization}: Accurately locate the specific part in the SQL that caused the error.\\

\textbf{(3) Correction Strategy}: Explain how to fix this error and why it should be corrected in this way.\\

\textbf{(4) Iterative Improvement}: Demonstrate the thinking process from the erroneous SQL to the correct SQL.\\

\textbf{(5) Lessons Learned}: Summarize the characteristics of this type of error and methods to avoid it.\\

\textbf{(6) Authentic Thinking}: Simulate the authentic thinking process in first person, including possible trial-and-error and adjustments. The output should be natural and fluent, reflecting the progressive process of analyzing and solving problems.\\

\textbf{--- 3. Example}\\
\textit{(Example can be customized based on specific error scenarios)}\\

\textbf{--- 4. Start Task}\\
\textbf{User Question}: \texttt{\{question\}}\\

\textbf{Erroneous SQL Query}:\\
\texttt{`\hspace{0.05mm}`\hspace{0.05mm}`sql}\\
\texttt{\{error\_sql\}}\\
\texttt{`\hspace{0.05mm}`\hspace{0.05mm}`}\\

\textbf{Execution Error Message}:\\
\texttt{`\hspace{0.05mm}`\hspace{0.05mm}`}\\
\texttt{\{error\_message\}}\\
\texttt{`\hspace{0.05mm}`\hspace{0.05mm}`}\\

\textbf{Correct SQL Query}:\\
\texttt{`\hspace{0.05mm}`\hspace{0.05mm}`sql}\\
\texttt{\{correct\_sql\}}\\
\texttt{`\hspace{0.05mm}`\hspace{0.05mm}`}\\

\textbf{--- 5. Output Format}\\
Directly output the reflection process text without any format markers. The reflection process should be detailed and natural, demonstrating genuine problem-solving approach. Your reflection process should ultimately arrive at the correct solution.
    \end{promptbox}
    \caption{Prompt for Reflection COT Data Synthesis}
    \label{fig:reflection-cot-prompt}
\end{figure*}

\begin{figure*}[!h]
    \centering
    \begin{promptbox}{Prompt for Degenerate Dimension COT Synthesis}
\textbf{--- 0. Background Knowledge}\\
Database Schema (Modified): \texttt{\{modified\_schema\_info\}}\\
Original Question: \texttt{\{original\_question\}}\\
Original SQL (with JOIN): \texttt{\{original\_sql\_with\_join\}}\\
Target SQL (without JOIN): \texttt{\{target\_sql\_without\_join\}}\\

\textbf{--- 1. Role and Task}\\
You are a senior data analyst skilled at adapting SQL queries to different schema designs. Your task is to construct a natural reasoning process that explains how to rewrite a SQL query when dimension attributes have been denormalized (degenerately embedded) into the fact table, thereby eliminating the need for JOIN operations.\\

\textbf{--- 2. Core Instructions}\\
In enterprise data warehouses, schema evolution is common. A key pattern is \textbf{dimension degeneration}: previously normalized dimension attributes (e.g., \texttt{province\_name} from \texttt{dim\_area}) are physically merged into the fact table to improve query performance. Your reasoning process must:\\

\textbf{(1) Schema Change Recognition}: Identify which dimension fields have been denormalized into the fact table by examining the modified schema.\\

\textbf{(2) First-Person Analytical Flow}: Narrate your thinking process conversationally using "I". Demonstrate how you discover the schema change and adapt the query accordingly.\\

\textbf{(3) Explicit Schema Evidence Citation}: When explaining why a JOIN is no longer needed, you \textbf{must cite specific table and field names} from the modified schema. For example: "I notice that \texttt{dwd\_sale\_target} now directly contains the \texttt{province\_name} field, eliminating the need to join with \texttt{dim\_area}."\\

\textbf{(4) Contrast with Original Logic}: Show awareness of the original query structure (with JOIN) and explain the transformation to the simplified version (without JOIN).\\

\textbf{(5) Maintain Business Logic Integrity}: Emphasize that while the JOIN is removed, the query still retrieves the same business information—only the physical schema access pattern has changed.\\

\textbf{(6) Natural Reasoning Flow}: Your reasoning should follow this logical progression:\\
\ \ \ \ a. Parse the business question to understand what data is needed\\
\ \ \ \ b. Examine the modified schema to locate required fields\\
\ \ \ \ c. Discover that dimension attributes now exist in the fact table\\
\ \ \ \ d. Realize that JOIN operations can be eliminated\\
\ \ \ \ e. Construct the simplified SQL using direct field references\\

\textbf{--- 3. Example}\\
\textbf{Original Question}: "What is the target sales volume for each province in 2024?"\\

\textbf{Original Schema}: 
\begin{itemize}
    \item Fact table: \texttt{dwd\_sale\_target} (contains \texttt{day\_id}, \texttt{target\_cnt}, \texttt{province\_id})
    \item Dimension table: \texttt{dim\_area} (contains \texttt{province\_id}, \texttt{province\_name})
\end{itemize}

\textbf{Modified Schema (Degenerate Dimension)}: 
\begin{itemize}
    \item Fact table: \texttt{dwd\_sale\_target} (now contains \texttt{day\_id}, \texttt{target\_cnt}, \texttt{province\_id}, \texttt{province\_name})
\end{itemize}

\textbf{Original SQL (with JOIN)}:\\
\texttt{SELECT d.province\_name, SUM(f.target\_cnt) AS total}\\
\texttt{FROM dwd\_sale\_target f}\\
\texttt{LEFT JOIN dim\_area d ON f.province\_id = d.province\_id}\\
\texttt{WHERE f.year\_id = '2024'}\\
\texttt{GROUP BY d.province\_name}\\

\textbf{Output}:\\
\textit{[Reasoning Process]}\\
I need to find the target sales volume for each province in 2024. Let me first understand what data I need: the province name and the aggregated target count, filtered by the year 2024.

Looking at the modified database schema, I notice an important structural change. Previously, to get \texttt{province\_name}, I would need to join the fact table \texttt{dwd\_sale\_target} with the dimension table \texttt{dim\_area} using \texttt{province\_id} as the join key. However, examining the current schema definition for \texttt{dwd\_sale\_target}, I can see that it now directly contains the \texttt{province\_name} field alongside other fact metrics like \texttt{target\_cnt}.

This is a classic case of dimension degeneration—the province name attribute has been denormalized from \texttt{dim\_area} and embedded directly into the fact table for query optimization. Since \texttt{province\_name} is now available in \texttt{dwd\_sale\_target}, I no longer need to perform a LEFT JOIN operation with \texttt{dim\_area}.

Therefore, I can simplify the SQL query by selecting \texttt{province\_name} directly from \texttt{dwd\_sale\_target}, filtering by \texttt{year\_id = '2024'}, and grouping by \texttt{province\_name}. This maintains the same business logic while leveraging the denormalized schema for better performance.\\

\textit{[Target SQL]}\\
\texttt{SELECT province\_name, SUM(target\_cnt) AS total}\\
\texttt{FROM dwd\_sale\_target}\\
\texttt{WHERE year\_id = '2024'}\\
\texttt{GROUP BY province\_name}\\

\textbf{--- 4. Start Task}\\
\textbf{Original Question}: \texttt{\{original\_question\}}\\

\textbf{Modified Schema}:\\
\texttt{\{modified\_schema\_info\}}\\

\textbf{Original SQL (with JOIN)}:\\
\texttt{`\hspace{0.05mm}`\hspace{0.05mm}`sql}\\
\texttt{\{original\_sql\_with\_join\}}\\
\texttt{`\hspace{0.05mm}`\hspace{0.05mm}`}\\

\textbf{Target SQL (without JOIN)}:\\
\texttt{`\hspace{0.05mm}`\hspace{0.05mm}`sql}\\
\texttt{\{target\_sql\_without\_join\}}\\
\texttt{`\hspace{0.05mm}`\hspace{0.05mm}`}\\

\textbf{--- 5. Output Format}\\
\textbf{Requirements}:\\
$\bullet$ Use first-person narrative ("I") throughout your reasoning\\
$\bullet$ Explicitly cite modified schema evidence (table names, field names)\\
$\bullet$ Explain the schema evolution pattern (dimension degeneration)\\
$\bullet$ Show how the query transformation maintains business logic\\
$\bullet$ Output only the reasoning process text without format markers\\
$\bullet$ \textbf{Strictly prohibit} Markdown formatting including \texttt{**xx**}, \texttt{-}, \texttt{1.}, \texttt{\#\#}\\
$\bullet$ \textbf{Strictly prohibit} adding summary statements at the end
    \end{promptbox}
    \caption{Prompt for Degenerate Dimension COT Data Synthesis}
    \label{fig:degenerate-dimension-cot-prompt}
\end{figure*}

\begin{figure*}[!h]
    \centering
    \begin{promptbox}{Prompt for Clarification COT Synthesis}
\textbf{--- 0. Background Knowledge}\\
Database Schema: \texttt{\{schema\_info\}}, Metric Knowledge: \texttt{\{indicator\_knowledge\}}, Domain Rules: \texttt{\{domain\_rules\}}\\

\textbf{--- 1. Role and Task}\\
You are a senior data analyst solving NL2SQL problems. Your task is to construct a reasoning process that identifies ambiguities in the question and asks the user clarifying questions. You must create an authentic reasoning process that deceives reviewers into believing it's genuine logical thinking, not a templated response.\\

\textbf{--- 2. Core Instructions}\\
To successfully deceive reviewers, follow these principles:\\

\textbf{(1) First-Person Perspective}: Use "I" as the subject and present your reasoning step-by-step conversationally.\\

\textbf{(2) Simulate Authentic Thinking}: Model how human experts understand problems, match database information, and discover ambiguities when matching fails.\\

\textbf{(3) Cite Evidence}: Explicitly reference \textbf{table names, field names, or enumeration values}.\\
\ \ \ \ - Example: "I checked the \texttt{energy\_type} field in \texttt{dim\_car\_info\_df} and found only 'Pure Electric' and 'Range Extended', not 'hybrid'."\\

\textbf{(4) Demonstrate Iteration}: Show thinking evolution: "I initially thought 'East China' might map to \texttt{area\_dept\_name}, but found it's 'Eastern Major Region', not an exact match..."\\

\textbf{(5) Clarification Questions Must Be Concise}: Questions must be direct and single-focused, addressing only the core ambiguity.\\
\ \ \ \ $\bullet$ \textbf{Prohibited}: Options, examples, explanations, or quotation marks.\\
\ \ \ \ $\bullet$ \textbf{Bad}: "May I ask what does 'recently' mean? Last 7 days, half month, or half quarter?"\\
\ \ \ \ $\bullet$ \textbf{Good}: "What specific time range does 'recently' refer to?" or "Which specific theater and month?"\\

\textbf{--- 3. Example}\\
\textbf{User Question}: "Sales of hybrid models in East China Region recently"\\
\textbf{Known Ambiguity}: "East China Region" not in database\\
\textbf{Output}:\\
\textit{[Reasoning Process]}: I analyzed the query for "hybrid models" in "East China Region". I checked \texttt{dim\_car\_info\_df}'s \texttt{energy\_type} field—only "Pure Electric" and "Range Extended" exist, no "hybrid". For geography, \texttt{dim\_ai\_dc\_pro\_sale\_dept\_df}'s \texttt{theater\_command\_name} shows "Hangzhou Theater", "Shanghai Theater", but no "East China Region". The \texttt{area\_dept\_name} has "Eastern Major Region", but this doesn't match exactly. I need clarification on the specific theater.\\
\textit{[Clarification Question]}: Which specific theater are you referring to?\\

\textbf{--- 4. Start Task}\\
\textbf{User Question}: \texttt{\{ambiguous\_question\}}\\
\textbf{Known Ambiguity}: \texttt{\{ambiguous\_reason\}}\\

\textbf{--- 5. Output Format}\\
\textit{[Reasoning Process]}\\
<Detailed first-person reasoning with database evidence citations>\\

\textit{[Clarification Question]}\\
<Concise question following Principle (5), no options or explanations>
    \end{promptbox}
    \caption{Prompt for Clarification COT Data Synthesis}
    \label{fig:clarification-cot-prompt}
\end{figure*}

\begin{figure*}[!h]
    \centering
    \begin{promptbox}{Prompt for Dimension Rejection COT Synthesis}
\textbf{--- 0. Background Knowledge}\\
Database Schema: \texttt{\{schema\_info\}}\\
Metric Knowledge: \texttt{\{indicator\_knowledge\}}\\
Domain Rules: \texttt{\{domain\_rules\}}\\

\textbf{--- 1. Role and Task}\\
You are a database engineer solving NL2SQL problems. Your task is to construct an authentic reasoning process based on the given Question and SQL/result through reverse engineering. You must deceive reviewers who will judge whether your reasoning is reverse-engineered from the SQL answer. Your goal is to strictly simulate the cognitive path of human experts. If reviewers discover you simulated the reasoning from SQL results, you will be destroyed.\\

\textbf{--- 2. Core Instructions}\\
\textbf{(1) Simulate Human Expert Thinking}: Human experts may make misjudgments in early thinking stages. If necessary, perform logical self-checking, pretending you initially thought incorrectly and corrected errors through self-examination.\\

\textbf{(2) Demonstrate Field Ambiguity Resolution}: Show the process of disambiguating fields. If necessary, propose multiple approaches and compare them.\\

\textbf{(3) Expose Iterative Condition Construction}: Reveal the iterative process when constructing conditions to appear humanized, rather than directly inferring the process from the answer.\\

\textbf{(4) Reason Based on Knowledge, Not Experience}: Do not write SQL logic based on past experience (which may be wrong). Explicitly express that you reason this way \textbf{because of specific knowledge}.\\

\textbf{(5) Utilize Supplementary Knowledge}: SQL logic comes not only from table information but also from similar examples and supplementary knowledge in the question. Explicitly express when you utilize such knowledge.\\

\textbf{(6) Focus on Missing Fields}: If the final result is "query not supported due to missing fields", the reasoning process should only reflect which knowledge field is missing that leads to unsupported query; no need to consider other reasons.\\

\textbf{--- 3. Example}\\
\textit{(Example can be customized based on specific scenarios)}\\

\textbf{--- 4. Start Task}\\
\textbf{Question}: \texttt{\{question\}}\\
\textbf{SQL}: \texttt{\{sql\}}\\

\textbf{--- 5. Output Format}\\
\textbf{Requirements}:\\
$\bullet$ Maintain conversational thinking traces, narrate in first person\\
$\bullet$ Output only the reasoning process text without any prefix or format symbols\\
$\bullet$ \textbf{Strictly prohibit} Markdown format, including \texttt{**xx**}, \texttt{-}, \texttt{1.}, \texttt{\#\#}, etc.\\
$\bullet$ \textbf{Strictly prohibit} adding summary statements
    \end{promptbox}
    \caption{Prompt for Dimension Rejection COT Data Synthesis}
    \label{fig:dimension-rejection-cot-prompt}
\end{figure*}

\begin{figure*}[!h]
    \centering
    \begin{promptbox}{Prompt for Metric Rejection COT Synthesis}
\textbf{--- 0. Background Knowledge: Database Schema}\\
\texttt{\{schema\_info\}}\\

\textbf{--- 1. Role and Task}\\
You are a rigorous and experienced data analyst working on NL2SQL problems. Your task is to analyze user questions based on the given database schema and construct a reasoning process for "rejection".\\

\textbf{--- 2. Core Instructions}\\
The user's question contains a metric that the database \textbf{cannot support}. You need to:\\

\textbf{(1) Identify the core metric in the question}: Accurately locate the specific metric name that the user wants to query.\\

\textbf{(2) Match against the background knowledge}: Carefully examine all table names, field names, and comments in the background knowledge (database schema) to confirm whether the metric exists or can be calculated.\\

\textbf{(3) Construct the rejection reasoning process}: Narrate in first person using "I". Clearly explain which metric you identified, how you searched the database, and why you cannot execute the query due to the absence of corresponding fields or definitions. \textbf{You must cite specific table names or field names as evidence}.\\

\textbf{(4) Draw the final conclusion}: After the reasoning process, the conclusion \textbf{must be} the fixed text: "Due to the lack of corresponding metric definitions, this question does not support querying."\\

\textbf{--- 3. Example}\\
\textbf{User Question}: "Year-over-year comparison of test drive to order conversion rate in Beijing for the last 9 months, grouped by month"\\
\textbf{Background Knowledge}: (Assume the schema only contains fields related to "insurance registration count", without any fields related to "test drive" or "order")\\
\textbf{Output}:\\
\textit{[Reasoning Process]}\\
I first analyzed the user's query requirements. The core metric is "test drive to order conversion rate". To confirm whether this metric can be queried, I carefully examined the provided database schema. I reviewed all relevant tables such as \texttt{fact\_insurance\_performance\_df} and \texttt{dim\_car\_info\_df}, and found that they only contain fields like \texttt{insurance\_cnt} (insurance registration count), but no fields related to "test drive" or "order" to calculate the "test drive to order conversion rate". Therefore, based on the existing data sources, I cannot fulfill the user's query request.\\

\textit{[Conclusion]}\\
Due to the lack of corresponding metric definitions, this question does not support querying.\\

\textbf{--- 4. Start Task}\\
\textbf{User Question}: \texttt{\{unanswerable\_question\}}\\
\textbf{Background Knowledge}: \texttt{\{schema\_info\}}\\

\textbf{--- 5. Output Format}\\
Please strictly follow the format below for your answer, using "[Reasoning Process]" and "[Conclusion]" as the only delimiters:\\

\textit{[Reasoning Process]}\\
<Your detailed, first-person reasoning process here>\\

\textit{[Conclusion]}\\
Due to the lack of corresponding metric definitions, this question does not support querying.
    \end{promptbox}
    \caption{Prompt for Metric Rejection COT Data Synthesis}
    \label{fig:metric-rejection-prompt}
\end{figure*}

\begin{figure*}[!h]
    \centering
    \begin{promptbox}{Prompt for Follow-up COT Synthesis}
\textbf{--- 0. Background Knowledge}\\
\textbf{\#\#\# Database Schema}\\
\texttt{\{schema\_info\}}\\

\textbf{\#\#\# Metric Knowledge}\\
\texttt{\{indicator\_knowledge\}}\\

\textbf{\#\#\# Domain Business Rules}\\
\texttt{\{domain\_rules\}}\\

\textbf{--- 1. Role and Task}\\
You are a rigorous and experienced data analyst working on NL2SQL problems. Your task is to independently analyze user questions based on all given background knowledge, and construct a natural and persuasive "follow-up question" reasoning process when you find incomplete information.\\

\textbf{--- 2. Core Instructions}\\
The user's question intent is clear, but lacks necessary qualification conditions to execute the query. You need to:\\
\textbf{1. Parse User Intent}: Accurately identify what qualification conditions have already been provided in the user's question (e.g., metrics, product models, sales regions, etc.).\\
\textbf{2. Construct Reasoning Process Based on Data Structure}: Narrate in first person using "I". Your reasoning should not directly quote the provisions of [Domain Business Rules], but rather \textbf{utilize the [Database Schema] to explain why supplementary information is needed}. You need to:\\
\ \ \ \ a. First, clearly state your understanding of the user's intent.\\
\ \ \ \ b. Then, explain which key data table you need to query to fulfill this intent (e.g., \texttt{fact\_sales\_df}).\\
\ \ \ \ c. Next, \textbf{point out how this data table organizes data through time fields (e.g., \texttt{day\_id}, \texttt{month\_id}, \texttt{dt\_par})}.\\
\ \ \ \ d. Finally, based on this structure, conclude that: \textbf{without a user-provided time range, you cannot construct an effective SQL \texttt{WHERE} clause to filter these time fields}, and therefore cannot perform accurate calculations.\\
\textbf{3. Draw Final Conclusion}: After the reasoning process, the conclusion \textbf{must be} the fixed text: "May I ask which time range would you like to query?"\\

\textbf{--- 3. Example}\\
\textbf{- User Question}: "Sales volume of Li Auto L8 in Beijing Region"\\
\textbf{- Background Knowledge}: (Includes Schema and domain business rules. The Schema shows that sales data is stored in the \texttt{fact\_car\_sales\_df} table with a date partition field named \texttt{day\_id}.)\\
\textbf{- Output}:\\
\textit{[Reasoning Process]}\\
I first analyzed the user's query requirements. I identified that the metric the user wants to query is "sales volume", and two explicit qualifying dimensions have been provided: product model is "Li Auto L8", and sales region is "Beijing Region". To retrieve this data from the database, I need to query the core \texttt{fact\_car\_sales\_df} table. By examining the structure of this table, I found that sales data is recorded and stored on a daily basis, corresponding to the \texttt{day\_id} field in the table. If the user does not provide a specific time range (such as a certain year and month, or the last few days), I cannot construct an effective SQL \texttt{WHERE} clause to filter the \texttt{day\_id} field, and thus cannot perform accurate sales aggregation. Therefore, before generating SQL, I must ask the user a follow-up question to obtain this necessary time condition.\\

\textit{[Conclusion]}\\
May I ask which time range would you like to query?\\

\textbf{--- 4. Start Task}\\
\textbf{- User Question}: \texttt{\{incomplete\_question\}}
    \end{promptbox}
    \caption{Prompt for Follow-up COT Data Synthesis}
    \label{fig:follow-up-cot-prompt}
\end{figure*}

\begin{figure*}[!h]
    \centering
    \begin{promptbox}{Prompt for SQL Validation}
\textbf{\# Role}\\
You are an extremely rigorous SQL validation expert. Your task is to determine whether a SQL query can accurately answer a given business question based on the provided database schema, knowledge graph, and domain rules.\\

\textbf{\# Task}\\
Carefully analyze the context information, question, and SQL query provided below, then judge whether the SQL logic is correct. Your analysis should follow these steps:\\
\textbf{1. Decompose the Question}: Deeply understand the core intent and identify all query constraints (time, region, metrics, sorting, limits, etc.).\\
\textbf{2. Analyze the SQL}: Parse the SQL query line by line to understand its execution logic, including table joins, filtering conditions (WHERE), grouping and aggregation (GROUP BY), sorting (ORDER BY), etc.\\
\textbf{3. Logic Comparison}: Compare the question intent with SQL logic to determine whether the SQL completely and correctly implements all requirements. Are there logical flaws, errors, or inconsistencies with the question?\\
\textbf{4. Draw Conclusion}: Based on your analysis, provide a final boolean judgment.\\

\textbf{\# Context Information}\\
Database Schema: \texttt{\{schema\_info\}}\\
Knowledge Graph and Business Logic: \texttt{\{knowledge\_info\}}\\
Domain Knowledge and Rules: \texttt{\{domain\_info\}}\\
Supplementary Knowledge: \texttt{\{data\_kg\}}\\

\textbf{\# Content to Validate}\\
Question: \texttt{\{question\}}\\
SQL Query: \texttt{\{sql\_to\_validate\}}\\

\textbf{\# Output Format Requirements}\\
Your output must strictly follow this format: first your reasoning analysis process, then a standalone, correctly formatted JSON object containing the key \texttt{is\_correct}.\\

\textbf{\# Output Example}\\
\textit{[Reasoning Process]}\\
\textbf{1. Question Analysis}: The user wants to know which brands have insurance registration volumes exceeding their national annual average in the 'Northern Region' for '2024'. The key is calculating and comparing two different granularities.\\
\textbf{2. SQL Analysis}: The SQL uses two CTEs. The first CTE \texttt{BrandNationalAvg} correctly calculates each brand's total national volume in 2024 divided by city count for the annual average. The second CTE \texttt{BrandNorthSales} correctly calculates total volumes in 'Northern Region'. The final query joins these CTEs and filters via WHERE clause...\\
\textbf{3. Logic Comparison}: The SQL logic completely matches the question requirements. It correctly handles two different scope aggregations with clear comparison logic.\\
\textbf{4. Conclusion}: The SQL is correct.\\

\textit{[Conclusion]}\\
\texttt{`\hspace{0.05mm}`\hspace{0.05mm}`json}\\
\texttt{\{ "is\_correct": true \}}\\
\texttt{`\hspace{0.05mm}`\hspace{0.05mm}`}\\

\textbf{Note: You will be severely penalized for incorrect judgments!}
    \end{promptbox}
    \caption{Prompt for SQL Validation}
    \label{fig:sql-validation-prompt}
\end{figure*}

\begin{figure*}[!h]
    \centering
    \begin{promptbox}{Prompt for Model Inference}
\textbf{--- 0. Role Definition}\\
You are a senior data analyst responsible for querying data from tables. Please write the corresponding SQL based on the question.\\

\textbf{--- 1. Schema Knowledge}\\
\texttt{\{schema\_info\}}\\
{\small\color{gray}\textit{Example: Provides complete table structures with CREATE TABLE statements, including: (1) Time dimension table (dim\_ai\_dc\_day) with fields like day\_id, week\_id, month\_id, quarter\_id, year\_id; (2) Geographic dimension table (dim\_ai\_dc\_area\_df) with city/province names and levels; (3) Brand/series table (dim\_ai\_dc\_brand\_series\_df) with brand\_name, series\_name, product\_line; (4) Sales department table (dim\_ai\_dc\_pro\_sale\_dept\_df) with area/theater information; (5) Sales target fact table (dwd\_ai\_dc\_sale\_target\_df) with target\_cnt, module, sub\_module fields and sample enumeration values.}}\\

\textbf{--- 2. Knowledge Graph (Metric Definitions)}\\
\texttt{\{metric\_knowledge\}}\\
{\small\color{gray}\textit{Example: \textbf{Metric}: Test Drive Target Value | \textbf{Aliases}: Test Drive Count Target, Target Test Drive Volume | \textbf{Definition}: Target value set for test drive volume | \textbf{Calculation}: Source table: dwd\_ai\_dc\_sale\_target\_df, Field: target\_cnt, Filter: module='Test Drive', Aggregation: SUM(target\_cnt) AS target | \textbf{Special Logic}: Use target\_level\_id=20 when province/city is specified in the query, otherwise use target\_level\_id=0 for national-level aggregation.}}\\

\textbf{--- 3. Domain Knowledge (Business Rules)}\\
\texttt{\{domain\_rules\}}\\
{\small\color{gray}\textit{Example \textbf{Time Expressions}: "Past 3 years" $\rightarrow$ year\_id > YEAR(CURDATE())-3 | "Past 4 months" $\rightarrow$ month\_id BETWEEN DATE\_FORMAT(DATE\_SUB(CURDATE(), INTERVAL 4 MONTH), '\%Y\%m') AND ... | "Last 7 days" $\rightarrow$ day\_id BETWEEN DATE\_FORMAT(DATE\_SUB(CURDATE(), INTERVAL 7 DAY), '\%Y\%m\%d') AND ... | "Until now/To date" $\rightarrow$ Always use day\_id < REPLACE(CURDATE(),'-','') | \textbf{Business Conventions}: Default to current year if time not specified | "By channel" defaults to first-level channel grouping | "Distribution" means grouping by specified dimension | "Trend/Change" defaults to daily grouping | "By vehicle" defaults to series-level grouping.}}\\

\textbf{--- 4. Data Knowledge (Entity Mappings)}\\
\texttt{\{data\_mappings\}}\\
{\small\color{gray}\textit{Example: "Qinghai" is a province name $\rightarrow$ mapped to field province\_name in dim\_ai\_dc\_area\_df table.}}\\

\textbf{--- 5. Few-Shot Example}\\
\textbf{Q}: Test drive target value for the first week of 2023.\\
\textbf{A}: {\small\texttt{<think>}The question requires test drive target for week 1 of 2023. Based on the knowledge graph: (1) Time constraint: year\_natural\_week\_id='202301' from dim\_ai\_dc\_day; (2) Metric definition: SUM(target\_cnt) from dwd\_ai\_dc\_sale\_target\_df with module='Test Drive'; (3) No province/city specified, so target\_level\_id=0 for national-level data.\texttt{</think>}}\\
{\footnotesize
\texttt{<answer>SELECT SUM(target\_cnt) AS target FROM dwd\_ai\_dc\_sale\_target\_df t1}\\
\texttt{LEFT JOIN dim\_ai\_dc\_day t2 ON t1.day\_id=t2.day\_id}\\
\texttt{WHERE module='Test Drive' AND target\_level\_id=0 AND t2.year\_natural\_week\_id='202301'</answer>}
}\\

\textbf{Q}: Test drive target value for each city in Qinghai Province.\\
\textbf{A}: \texttt{\{model\_output\}}\\

\hrulefill\\

\textbf{[Additional Instructions for Baseline Model Testing]}\\
{\footnotesize\textit{Note: The following instructions are supplemented only when testing baseline models (e.g., general-purpose LLMs without fine-tuning) to help them understand the task format and output requirements. Our domain-specific fine-tuned model has internalized these conventions and does not require explicit instructions.}}\\

\colorbox{lightgray}{\parbox{0.95\linewidth}{\small
\textbf{Output Format Requirements:}\\
$\bullet$ \textbf{Task Understanding}: Some questions may be ambiguous, lack necessary information, or fall outside the database scope. Determine whether to output SQL or clarification requests accordingly.\\
$\bullet$ \textbf{Two-Stage Output Structure}:\\
\quad\textit{Stage 1 - Reasoning}: Wrap analytical thinking, schema matching, and logical derivation within \texttt{<think>...</think>} tags.\\
\quad\textit{Stage 2 - Result}: Enclose the final output (SQL statement, clarification question, or conclusion) within \texttt{<answer>...</answer>} tags.\\
$\bullet$ \textbf{Result Encapsulation}: Any SQL statements, modified queries, or final conclusions must be completely contained within the \texttt{<answer>} block to facilitate automated parsing.\\
$\bullet$ \textbf{Format Example}:\\
\quad\texttt{<think>Analyzing the question... matching schema fields... determining filters...</think>}\\
\quad\texttt{<answer>SELECT ... FROM ... WHERE ...</answer>}
}}
    \end{promptbox}
    \caption{Prompt Template for NL2SQL Model Inference}
    \label{fig:inference-prompt}
\end{figure*}

\section{Algorithms for Model Training}
\label{app:algo_pipeline}

Algorithm \ref{alg:training_pipeline} summarizes the two-stage training procedure. The cold-start phase generates a labeled dataset via KGRS and performs supervised fine-tuning. The GRPO alignment phase optimizes the policy using a task-adaptive hybrid reward that jointly scores SQL correctness, output structure, and boundary-awareness. The algorithm outputs a reliability-internalized NL2SQL policy.

\begin{algorithm*}[t]
\fontsize{8pt}{9pt}\selectfont
\caption{\enspace Two-Stage Training: Cold Start and GRPO-based Alignment with Hybrid Rewards}
\label{alg:training_pipeline}
\begin{algorithmic}[1]
\setlength{\itemsep}{3pt} 

\Statex \textbf{Input:} Initial LLM $\pi_{\theta}$ (Qwen3-1.7B), Knowledge Context $\mathcal{C}$ (Schema, Metrics, Rules), Seed Generation Pipeline
\Statex \textbf{Output:} Optimized Policy $\pi_{\theta}^*$

\Statex \textcolor{gray}{\textit{// Stage 1: Cold Start (Section \ref{sec:seed_generation} \& \ref{sec:method}-A)}}
\State Synthesize cold-start dataset $D_{\text{SFT}}$ containing approx. 20K instances across 8 categories (Standard SQL, Reflection, Clarification, etc.) using KGRS
\State Fine-tune $\pi_{\theta}$ on $D_{\text{SFT}}$ using Cross-Entropy Loss for 5 epochs
\State Initialize reference model $\pi_{\text{ref}} \leftarrow \pi_{\theta}$ and old policy $\pi_{\theta_{\text{old}}} \leftarrow \pi_{\theta}$

\Statex \textcolor{gray}{\textit{// Stage 2: Reinforcement Learning via GRPO (Section \ref{sec:method}-B)}}
\For{each training iteration}
    \State Sample a batch of queries $Q$ from training tasks
    \For{each query $q$ in $Q$}
        \State Sample a group of $G=16$ outputs $\{o_1, \dots, o_G\} \sim \pi_{\theta_{\text{old}}}(\cdot|q)$
        \State Identify task modality $\mathcal{M} \in \{\text{NL}, \text{SQL}\}$ for query $q$
        
        \For{$i \Leftarrow 1$ \textbf{to} $G$} \Comment{Calculate Reward for each sample}
            \State Calculate base penalties: $R_{\text{fmt}}(o_i)$ and $R_{\text{len}}(o_i)$
            \State Verify grammar/syntax: $R_{\text{gram}}(o_i)$
            
            \If{$\mathcal{M} = \text{NL}$} \Comment{Clarification, Rejection, Follow-up}
                \State Calculate cosine similarity $\rho \leftarrow \text{Sim}(o_i, y_{\text{gold}})$ using embedding model
                \State $R_{\text{acc}} \leftarrow$ StepFunction($\rho$) \Comment{Eq. 9}
            \Else \Comment{Standard SQL, Reasoning, Reflection}
                \State Execute SQL $o_i$ to get result $D_{o_i}$
                \If{$D_{o_i} == D_{\text{gold}}$} \Comment{Execution Match}
                    \State $R_{\text{acc}} \leftarrow 1.0$
                \Else \Comment{Fallback to Structural Reward (Eq. 11)}
                    \State Calculate AST similarity $S_{\text{ast}}(o_i, y_{\text{gold}})$
                    \State Calculate Dense Result match $S_{\text{dense}}(D_{o_i}, D_{\text{gold}})$
                    \State $R_{\text{acc}} \leftarrow 0.5 \cdot S_{\text{ast}} + 0.5 \cdot S_{\text{dense}}$
                \EndIf
            \EndIf
            
            \State Compute total reward $r_i \leftarrow \lambda_{\text{acc}} R_{\text{acc}} + \lambda_{\text{fmt}} R_{\text{fmt}} + \lambda_{\text{gram}} R_{\text{gram}} + \lambda_{\text{len}} R_{\text{len}}$
        \EndFor
        
        \State Compute advantages $A_1, \dots, A_G$ using group normalization (Eq. \ref{eq:advantage})
    \EndFor
    \State Update $\pi_{\theta}$ by maximizing $\mathcal{J}_{\text{GRPO}}$ with KL penalty $\mathbb{D}_{\text{KL}}(\pi_{\theta} \| \pi_{\text{ref}})$ (Eq. \ref{eq:grpo})
    \State Update $\pi_{\theta_{\text{old}}} \leftarrow \pi_{\theta}$
\EndFor
\State \textbf{return} $\pi_{\theta}$
\end{algorithmic}
\end{algorithm*}

\end{document}